# Evolution with Drifting Targets

**Varun Kanade**[*]
Harvard University
vkanade@fas.harvard.edu

**Leslie G. Valiant**[*]
Harvard University
valiant@seas.harvard.edu

**Jennifer Wortman Vaughan**[†]
Harvard University
jenn@seas.harvard.edu

## Abstract

We consider the question of the stability of evolutionary algorithms to gradual changes, or *drift*, in the target concept. We define an algorithm to be resistant to drift if, for some inverse polynomial drift rate in the target function, it converges to accuracy $1 - \epsilon$ with polynomial resources, and then stays within that accuracy indefinitely, except with probability $\epsilon$ at any one time. We show that every evolution algorithm, in the sense of Valiant [19], can be converted using the Correlational Query technique of Feldman [9], into such a drift resistant algorithm. For certain evolutionary algorithms, such as for Boolean conjunctions, we give bounds on the rates of drift that they can resist. We develop some new evolution algorithms that are resistant to significant drift. In particular, we give an algorithm for evolving linear separators over the spherically symmetric distribution that is resistant to a drift rate of $O(\epsilon/n)$, and another algorithm over the more general product normal distributions that resists a smaller drift rate.

The above translation result can be also interpreted as one on the robustness of the notion of evolvability itself under changes of definition. As a second result in that direction we show that every evolution algorithm can be converted to a quasi-monotonic one that can evolve from any starting point without the performance ever dipping significantly below that of the starting point. This permits the somewhat unnatural feature of arbitrary performance degradations to be removed from several known robustness translations.

## 1 Overview

The evolvability model introduced by Valiant [19] was designed to provide a quantitative theory for studying mechanisms that can evolve in populations of realistic size, in a reasonable number of generations through the Darwinian process of variation and selection. It models evolving mechanisms as functions of many arguments, where the value of a function represents the outcome of the mechanism, and the arguments the controlling factors. For example, the function might determine the expression level of a particular protein given the expression levels of related proteins. Evolution is then modeled as a restricted form of learning from examples, in which the learner observes only the empirical performance of a set of functions that are feasible variants of the current function. The *performance* of a function is defined as its correlation with the *ideal* function, which specifies for every possible circumstance the behavior that is most beneficial in the current environment for the evolving entity.

The evolution process consists of repeated applications of a random variation step followed by a selection step. In the variation step of round $i$, a polynomial number of variants of the algorithm's current hypothesis $r_i$ are generated, and their performance empirically tested. In the selection step, one of the variants with high performance is chosen as $r_{i+1}$. An algorithm therefore consists of both a procedure for describing possible variants and as well as a selection mechanism for choosing among the variants. The algorithm succeeds if it produces a hypothesis with performance close to the ideal function using only a polynomial amount of resources (in terms of number of generations and population size).

---

[*]This work was supported in part by NSF-CCF-04-27129.

[†]Vaughan is supported by NSF under grant CNS-0937060 to the CRA for the CIFellows Project. Any opinions, findings, conclusions, or recommendations expressed in this material are those of the authors alone.

The basic model as defined in Valiant [19] is concerned with the evolution of Boolean functions using representations that are randomized Boolean functions. This has been shown by Feldman [10] to be a highly robust class under variations in definition, as is necessary for any computational model that aims to capture the capabilities and limitations of a natural phenomenon. This model has also been extended to allow for representations with real number values, in which case a range of models arise that differ according to whether the quadratic loss or some other metric is used in evaluating performance [17, 10]. Our interest here remains with the original Boolean model, which is invariant under changes of this metric.

In this paper we consider the issue of stability of an evolution algorithm to gradual changes, or *drift*, in the target or ideal function. Such stability is a desirable property of evolution algorithms that is not explicitly captured in the original definition. We present two main results in this paper. First, for specific evolution algorithms we quantify how resistant they are to drift. Second, we show that evolutionary algorithms can be transformed to stable ones, showing that the evolutionary model is robust also under modifications that require resistance to drift.

The issue of resistance to drift has been discussed informally before in the context of evolution algorithms that are monotone in the sense that their performance is increasing, or at least non-decreasing, at every stage [17, 10]. We shall therefore start by distinguishing among three notions of monotonicity in terms of properties that need to hold with high probability: (i) quasi-monotonic, where for any $\epsilon$ the performance never goes more than $\epsilon$ below that of the starting hypothesis $r_0$, (ii) monotonic, where the performance never goes below that of $r_0$, and (iii) strictly monotonic, where performance increases by at least an inverse polynomial amount at each step. Definition (ii) is essentially Feldman's [10] and definition (iii) is implicit in Michael [17].

We define a notion of an evolution algorithm being stable to drift in the sense that for some inverse polynomial amount of drift, using only polynomial resources, the algorithm will converge to performance $1 - \epsilon$, and will stay with such high performance in perpetuity in the sense that at every subsequent time, except with probability $\epsilon$, its performance will be at least $1 - \epsilon$.

As our main result demonstrating the robustness of the evolutionary model itself, we show, through the simulation of query learning algorithms [9], that for every distribution $D$, every function class that is evolvable in the original definition, is also evolvable by an algorithm that is both (i) quasi-monotonic, and (ii) stable to some inverse polynomial amount of drift. While the definitions allow any small enough inverse polynomial drift rate, they require good performance in perpetuity, and with the same representation class for all $\epsilon$. Some technical complications arise as a result of the latter two requirements.

As a vehicle for studying the stability of specific algorithms, we show that there are natural evolutionary algorithms for linear separators over symmetric distributions and over the more general product normal distributions. Further we formulate a general result that states that for any strictly monotonic evolution algorithm, where the increase in performance at every step is defined by an inverse polynomial $b$, one can determine upper bounds on the polynomial parameters of the evolution algorithm, namely those that bound the generation numbers, population sizes, and sample sizes, and also a lower bound on the drift that can be resisted. We illustrate the usefulness of this formulation by applying it to show that our algorithms for linear separators can resist a significant amount of drift. We also apply it to existing algorithms for evolving conjunctions over the uniform distribution, with or without negations. We note that the advantages of evolution algorithms that use natural representations, over those obtained through simulations of query learning algorithms, may be quantified in terms of how moderate the degrees are of the polynomials that bound the generation number, population size, sample size and (inverse) drift rate of these algorithms. These results appear in Sections 6 and 7 and may be read independently of Section 5.

All omitted details and proofs appear in the appendix.

## 2 The Computational Model of Evolution

In this section, we provide an overview of the original computational model of evolution (Valiant [19], where further details can be found). Many of these notions will be familiar to readers who are acquainted with the PAC model of learning [18].

### 2.1 Basic Definitions

Let $\mathcal{X}$ be a space of examples. A *concept class* $\mathcal{C}$ over $\mathcal{X}$ is a set of functions mapping elements in $\mathcal{X}$ to $\{-1, 1\}$. A *representation class* $\mathcal{R}$ over $\mathcal{X}$ consists of a set of (possibly randomized) functions from $\mathcal{X}$ to $\{-1, 1\}$ described in a particular language. Throughout this paper, we think of $\mathcal{C}$ as the class of functions from which the ideal target $f$ is selected, and $\mathcal{R}$ as a class of representations from which the evolutionary algorithm chooses an $r$ to approximate $f$. We consider only classes of representations that can be evaluated efficiently, that is, classes $\mathcal{R}$ such that for any $r \in \mathcal{R}$ and any $x \in \mathcal{X}$, $r(x)$ can be evaluated in time polynomial in the size of $x$.

We associate a *complexity parameter* $n$ with $\mathcal{X}$, $\mathcal{C}$, and $\mathcal{R}$. This parameter indicates the number of dimensions of each element in the domain. For example, we might define $\mathcal{X}_n$ to be $\{-1, 1\}^n$, $\mathcal{C}_n$ to be the class of monotone conjunctions over $n$ variables, and $\mathcal{R}_n$ to be the class of monotone conjunctions over $n$ variables with each conjunction represented as a list of variables. Then $\mathcal{C} = \{\mathcal{C}_n\}_{n=1}^{\infty}$ and $\mathcal{R} = \{\mathcal{R}_n\}_{n=1}^{\infty}$ are really ensembles of classes.[1] Many of our results depend on this complexity parameter $n$. However, we drop the subscripts when the meaning is clear from context.

The *performance* of a representation $r$ with respect to the ideal target $f$ is measured with respect to a distribution $\mathcal{D}$ over examples. This distribution represents the relative frequency with which the organism faces each set of conditions in $\mathcal{X}$. Formally, for any pair of functions $f : \mathcal{X} \to \{-1, 1\}$, $r : \mathcal{X} \to \{-1, 1\}$, and distribution $\mathcal{D}$ over $\mathcal{X}$, we define the *performance* of $r$ with respect to $f$ as

$$\texttt{Perf}_f(r, \mathcal{D}) = \mathrm{E}_{x \sim \mathcal{D}}[f(x)r(x)] = 1 - 2\texttt{err}_{\mathcal{D}}(f, r) \; ,$$

where $\texttt{err}_{\mathcal{D}}(f, r) = \Pr_{x \sim \mathcal{D}}(f(x) \neq r(x))$ is the 0/1 error between $f$ and $r$. The performance thus measures the correlation between $f$ and $r$ and is always between $-1$ and $1$.

A new mutation is selected after each round of variation based in part on the observed fitness of the variants, i.e., their empirical correlations with the target on a polynomial number of examples. Formally, the *empirical performance* of $r$ with respect to $f$ on a set of examples $x_1, \cdots, x_s$ chosen independently according to $\mathcal{D}$ is a random variable defined as $(1/s) \sum_{i=1}^{s} f(x_i)r(x_i)$.

We denote by $\epsilon$ an *accuracy parameter* specifying how close to the ideal target a representation must be to be considered good. A representation $r$ is a good approximation of $f$ if $\texttt{Perf}_f(r, \mathcal{D}) \geq 1 - \epsilon$ (or equivalently, if $\texttt{err}_{\mathcal{D}}(f, r) \leq \epsilon/2$). We allow the evolution algorithm to use resources that are polynomial in both $1/\epsilon$ and the dimension $n$.

## 2.2 Model of Variation and Selection

An *evolutionary algorithm* $\mathcal{E}$ determines at each round $i$ which set of mutations of the algorithm's current hypothesis $r_{i-1}$ should be evaluated as candidates for $r_i$, and how the selection will be made. The algorithm $\mathcal{E} = (\mathcal{R}, \texttt{Neigh}, \mu, t, s)$ is specified by the following set of components:

- The *representation class* $\mathcal{R} = \{\mathcal{R}_n\}_{n=1}^{\infty}$ specifies the space of representations over $\mathcal{X}$ from which the algorithm may choose functions $r$ to approximate the target $f$.

- The (possibly randomized) function $\texttt{Neigh}(r, \epsilon)$ specifies for each $r \in \mathcal{R}_n$ the set of representations $r' \in \mathcal{R}_n$ into which $r$ can randomly mutate. This set of representations is referred to as the *neighborhood* of $r$. For all $r$ and $\epsilon$, it is required that $r \in \texttt{Neigh}(r, \epsilon)$ and that the size of the neighborhood is upper bounded by a polynomial.

- The function $\mu(r, r', \epsilon)$ specifies for each $r \in \mathcal{R}_n$ and each $r' \in \texttt{Neigh}(r, \epsilon)$ the probability that $r$ mutates into $r'$. It is required that for all $r$ and $\epsilon$, for all $r' \in \texttt{Neigh}(r, \epsilon)$, $\mu(r, r', \epsilon) \geq 1/p(n, 1/\epsilon)$ for a polynomial $p$.

- The function $t(r, \epsilon)$, referred to as the *tolerance* of $\mathcal{E}$, determines the difference in performance that a mutation in the neighborhood of $r$ must exhibit in order to be considered a "beneficial", "neutral", or "deleterious" mutation. The tolerance is required to be bounded from above and below, for all representations $r$, by a pair of inverse polynomials in $n$ and $1/\epsilon$.

- Finally, the function $s(r, \epsilon)$, referred to as the *sample size*, determines the number of examples used to evaluate the empirical performance of each $r' \in \texttt{Neigh}(r, \epsilon)$. The sample size must also be polynomial in $n$ and $1/\epsilon$.

The functions $\texttt{Neigh}$, $\mu$, $t$, and $s$ must all be computable in time polynomial in $n$ and $1/\epsilon$.

We are now ready to describe a single round of the evolution process. For any ideal target $f \in \mathcal{C}$, distribution $\mathcal{D}$, evolutionary algorithm $\mathcal{E} = (\mathcal{R}, \texttt{Neigh}, \mu, t, s)$, accuracy parameter $\epsilon$, and representation $r_{i-1}$, the *mutator* $\mathrm{M}(f, \mathcal{D}, \mathcal{E}, \epsilon, r_{i-1})$ returns a random mutation $r_i \in \texttt{Neigh}(r_{i-1}, \epsilon)$ using the following selection procedure. First, for each $r \in \texttt{Neigh}(r_{i-1}, \epsilon)$, the mutator computes the empirical performance of $r$ with respect to $f$ on a sample of size $s$.[2] Call this $v(r)$. Let

$$\texttt{Bene} = \{r \mid r \in \texttt{Neigh}(r_{i-1}, \epsilon), \; v(r) \geq v(r_{i-1}) + t(r_{i-1}, \epsilon)\}$$

---
[1] As in the PAC model, $n$ should additionally upper bound the size of representation of the function to be learned, but for brevity we shall omit this aspect here.

[2] We assume a single sample is used to evaluate the performance of all neighbors and $r_{i-1}$, but one could interpret the model as using independent samples for each representation. This would not change our results.

be the set of "beneficial" mutations and

$$\texttt{Neut} = \{r \mid r \in \texttt{Neigh}(r_{i-1}, \epsilon),\ |v(r) - v(r_{i-1})| < t(r_{i-1}, \epsilon)\}$$

be the set of "neutral" mutations. If at least one beneficial mutation exists, then a mutation $r$ is chosen from Bene as the survivor $r_i$ with relative probability $\mu(r_{i-1}, r, \epsilon)$. If no beneficial mutation exists, then a mutation $r$ is chosen from Neut as the survivor $r_i$, again with probability proportional to $\mu(r_{i-1}, r, \epsilon)$. Notice that, by definition, $r_{i-1}$ is always a member of Neut, and hence a neutral mutation is guaranteed to exist.

### 2.3 Putting It All Together

A concept class $\mathcal{C}$ is said to be evolvable by algorithm $\mathcal{E}$ over distribution $\mathcal{D}$ if for every target $f \in \mathcal{C}$, starting at any $r_0 \in \mathcal{R}$, the sequence of mutations defined by $\mathcal{E}$ converges in polynomial time to a representation $r$ whose performance with respect to $f$ is close to 1. This is formalized as follows.

**Definition 1 (Evolvability [19])** *For a concept class $\mathcal{C}$, distribution $\mathcal{D}$, and evolutionary algorithm $\mathcal{E} = (\mathcal{R}, \texttt{Neigh}, \mu, t, s)$, we say that $\mathcal{C}$ is evolvable over $\mathcal{D}$ by $\mathcal{E}$ if there exists a polynomial $g(n, 1/\epsilon)$ such that for every $n \in \mathbb{N}$, $f \in \mathcal{C}_n$, $r_0 \in \mathcal{R}_n$, and $\epsilon > 0$, with probability at least $1 - \epsilon$, a sequence $r_0, r_1, r_2, \cdots$ generated by setting $r_i = M(f, \mathcal{D}, \mathcal{E}, \epsilon, r_{i-1})$ for all $i$ satisfies $\texttt{Perf}_f(r_{g(n, 1/\epsilon)}, \mathcal{D}) \geq 1 - \epsilon$.*

We say that the class $\mathcal{C}$ is *evolvable over $\mathcal{D}$* if there exists a valid evolution algorithm $\mathcal{E} = (\mathcal{R}, \texttt{Neigh}, \mu, t, s)$ such that $\mathcal{C}$ is evolvable over $\mathcal{D}$ by $\mathcal{E}$. The polynomial $g(n, 1/\epsilon)$, referred to as the *generation polynomial*, is an upper bound on the number of generations required for the evolution process to converge. If the above definition holds only for a particular value (or set of values) for $r_0$, then we say that $\mathcal{C}$ is evolvable *with initialization*.

### 2.4 Alternative Models

Various alternative formulations of the basic computational model of evolution described here have been studied. Many have been proved equivalent to the basic model in the sense that any concept class $\mathcal{C}$ evolvable in the basic model is evolvable in the alternative model and vice versa. Here we briefly discuss some of the variations that have been considered.

The performance measure $\texttt{Perf}_f(r, \mathcal{D})$ is defined in terms of the 0/1 loss. Alternative performance measures based on squared loss or other loss functions have been studied in the context of evolution [10, 11, 17]. However, these alternative measures are identical to the original when $f$ and $r$ are (possibly randomized) binary functions, as we have assumed. (When the model is extended to allow real-valued function output, evolvability with a performance measure based on any nonlinear loss function is strictly more powerful than evolvability with the standard correlation-based performance measure [10]. We do not consider that extension in this work.)

Alternate rules for determining how a mutation is selected have also been considered. In particular, Feldman [10] showed that evolvability using a selection rule that always chooses among the mutations with the highest or near highest empirical performance in the neighborhood is equivalent to evolvability with the original selection rule based on the classes Bene and Neut. He also discussed the performance of "smooth" selection rules, in which the probability of a given mutation surviving is a smooth function of its original frequency and the performance of mutations in the neighborhood.

Finally, Feldman [9, 10] showed that *fixed-tolerance* evolvability, in which the tolerance $t$ is a function of only $n$ and $1/\epsilon$ but not the representation $r_{i-1}$, is equivalent to the basic model.

## 3 Notions of Monotonicity

Feldman [10, 11] introduced the notion of monotonic evolution in the computational model described above. His notion of monotonicity, restated here in Definition 2, requires that with high probability, the performance of the current representation $r_i$ never drops below the performance of the initial representation $r_0$ during the evolution process.

**Definition 2 (Monotonic Evolution)** *An evolution algorithm $\mathcal{E}$ monotonically evolves a class $\mathcal{C}$ over a distribution $\mathcal{D}$ if $\mathcal{E}$ evolves $\mathcal{C}$ over $\mathcal{D}$ and with probability at least $1 - \epsilon$, for all $i \leq g(n, 1/\epsilon)$, $\texttt{Perf}_f(r_i, \mathcal{D}) \geq \texttt{Perf}_f(r_0, \mathcal{D})$, where $g(n, 1/\epsilon)$ and $r_0, r_1, \cdots$ are defined as in Definition 1.*

When explicit initialization of the starting representation $r_0$ is prohibited, this is equivalent to requiring that $\texttt{Perf}_f(r_i, \mathcal{D}) \geq \texttt{Perf}_f(r_{i-1}, \mathcal{D})$ for all $i \leq g(n, 1/\epsilon)$. In other words, it is equivalent to requiring that with high probability, performance never decreases during the evolution process.

(Feldman showed that if representations may produce real-valued output and an alternate performance measure based on squared loss in considered, then any class $\mathcal{C}$ that is efficiently SQ learnable over a known, efficiently samplable distribution $\mathcal{D}$ is monotonically evolvable over $\mathcal{D}$.)

A stronger notion of monotonicity was used by Michael [17], who, in the context of real-valued representations and quadratic loss functions, developed an evolution algorithm for learning 1-decision lists in which only beneficial mutations are allowed. In this spirit, we define the notion of *strict* monotonic evolution, which requires a significant (inverse polynomial) performance increase at every round of evolution until a representation with sufficiently high performance is found.

**Definition 3 (Strict Monotonic Evolution)** *An evolution algorithm $\mathcal{E}$ strictly monotonically evolves a class $\mathcal{C}$ over a distribution $\mathcal{D}$ if $\mathcal{E}$ evolves $\mathcal{C}$ over $\mathcal{D}$ and, for a polynomial $m$, with probability at least $1 - \epsilon$, for all $i \leq g(n, 1/\epsilon)$, either $\text{Perf}_f(r_{i-1}, \mathcal{D}) \geq 1 - \epsilon$ or $\text{Perf}_f(r_i, \mathcal{D}) \geq \text{Perf}_f(r_{i-1}, \mathcal{D}) + 1/m(n, 1/\epsilon)$, where $g(n, 1/\epsilon)$ and $r_0, r_1, \cdots$ are defined as in Definition 1.*

Below we show that a class $\mathcal{C}$ is strictly monotonically evolvable over a distribution $\mathcal{D}$ using representation class $\mathcal{R}$ if and only if it is possible to define a neighborhood function satisfying the property that for any $r \in \mathcal{R}$ and $f \in \mathcal{C}$, if $\text{Perf}_f(r, \mathcal{D})$ is not already near optimal, there exists a neighbor $r'$ of $r$ such that $r'$ has a noticeable (again, inverse polynomial) performance improvement over $r$. We call such a neighborhood function *strictly beneficial*. The idea of strictly beneficial neighborhood functions plays an important role in developing our results in Sections 6 and 7. Feldman [11] uses a similar notion to show monotonic evolution under square loss.

**Definition 4 (Strictly Beneficial Neighborhood Function)** *For a concept class $\mathcal{C}$, distribution $\mathcal{D}$, and representation class $\mathcal{R}$, we say that a (possibly randomized) function Neigh is a strictly beneficial neighborhood function if the size of $\text{Neigh}(r, \epsilon)$ is upper bounded by a polynomial $p(n, 1/\epsilon)$, and there exists a polynomial $b(n, 1/\epsilon)$ such that for every $n \in \mathbb{N}$, $f \in \mathcal{C}_n$, $r \in \mathcal{R}_n$, and $\epsilon > 0$, if $\text{Perf}_f(r, \mathcal{D}) < 1 - \epsilon/2$, then there exists a $r' \in \text{Neigh}(r, \epsilon)$ such that $\text{Perf}_f(r', \mathcal{D}) \geq \text{Perf}_f(r, \mathcal{D}) + 1/b(n, 1/\epsilon)$. We refer to $b(n, 1/\epsilon)$ as the* benefit polynomial.

**Lemma 5** *For any concept class $\mathcal{C}$, distribution $\mathcal{D}$, and representation class $\mathcal{R}$, if Neigh is a strictly beneficial neighborhood function for $\mathcal{C}$, $\mathcal{D}$, and $\mathcal{R}$, then there exist valid functions $\mu, t,$ and $s$ such that $\mathcal{C}$ is strictly monotonically evolvable over $\mathcal{D}$ by $\mathcal{E} = (\mathcal{R}, \text{Neigh}, \mu, t, s)$. If a concept class $\mathcal{C}$ is strictly monotonically evolvable over $\mathcal{D}$ by $\mathcal{E} = (\mathcal{R}, \text{Neigh}, \mu, t, s)$, then Neigh is a strictly beneficial neighborhood function for $\mathcal{C}$, $\mathcal{D}$, and $\mathcal{R}$.*

The proof of the second half of the lemma is immediate; the definition of strictly monotonic evolvability requires that for any initial representation $r_0 \in \mathcal{R}$, with high probability either $\text{Perf}_f(r_0, \mathcal{D}) \geq 1 - \epsilon/2$ or $\text{Perf}_f(r_1, \mathcal{D}) \geq \text{Perf}_f(r_0, \mathcal{D}) + 1/m(n, 2/\epsilon)$ for a polynomial $m$. Thus if $\text{Perf}_f(r_0, \mathcal{D}) < 1 - \epsilon/2$ there must exist an $r_1$ in the neighborhood of $r_0$ such that $\text{Perf}_f(r_1, \mathcal{D}) \geq \text{Perf}_f(r_0, \mathcal{D}) + 1/m(n, 2/\epsilon)$. The key idea behind the proof of the first half is to show that it is possible to set the tolerance $t(r, \epsilon)$ in such a way that with high probability, Bene is never empty and there is never a representation in Bene with performance too much worse than that of the beneficial mutation guaranteed by the definition of the strictly beneficial neighborhood function. This implies that the mutation algorithm is guaranteed to choose a new representation with a significant increase in performance at each round.

Finally, we define quasi-monotonic evolution. This is similar to the monotonic evolution, except that the performance is allowed to go slightly below that of $r_0$. In Section 5.7, we show that this notion can be made universal, in the sense that every evolvable class is also evolvable quasi-monotonically.

**Definition 6 (Quasi-Monotonic Evolution)** *An evolution algorithm quasi-monotonically evolves a class $\mathcal{C}$ over $\mathcal{D}$ if $\mathcal{E}$ evolves $\mathcal{C}$ over $\mathcal{D}$ and with probability at least $1 - \epsilon$, for all $i \leq g(n, 1/\epsilon)$, $\text{Perf}_f(r_i, \mathcal{D}) \geq \text{Perf}_f(r_0, \mathcal{D}) - \epsilon$, where $g(n, 1/\epsilon)$ and $r_0, r_1, \cdots$ are defined as in Definition 1.*

## 4 Resistance to Drift

There are many ways one could choose to formalize the notion of drift resistance. Our formalization is closely related to ideas from the work on tracking drifting concepts in the computational learning literature. The first models of concept drift were proposed around the same time by Helmbold and Long [12] and Kuh et al. [16]. In both of these models, at each time $i$, an input point $x_i$ is drawn from a fixed but unknown distribution $\mathcal{D}$ and labeled by a target function $f_i \in \mathcal{C}$. It is assumed that the error of $f_i$ with respect to $f_{i-1}$ on $\mathcal{D}$ is less than a fixed value $\Delta$. Helmbold and Long [12] showed that a simple algorithm that chooses a concept to (approximately) minimize error over recent time steps achieves an average error of $\tilde{O}(\sqrt{\Delta d})$ where $d$ is the VC dimension of $\mathcal{C}$.[3] More general models of drift have also been proposed [2, 3].

Let $f_i \in \mathcal{C}$ denote the ideal function on round $i$ of the evolution process. Following Helmbold and Long [12], we make the assumption that for all $i$, $\text{err}_\mathcal{D}(f_{i-1}, f_i) \leq \Delta$ for some value $\Delta$. This is equivalent to assuming that $\text{Perf}_{f_{i-1}}(f_i, \mathcal{D}) \geq 1 - 2\Delta$. Call a sequence of functions satisfying this condition a $\Delta$-drifting sequence. We make no other assumptions on the sequence of ideal functions.

**Definition 7 (Evolvability with Drifting Targets)** *For a concept class $\mathcal{C}$, distribution $\mathcal{D}$, and evolution algorithm $\mathcal{E} = (\mathcal{R}, \texttt{Neigh}, \mu, t, s)$, we say that $\mathcal{C}$ is evolvable with drifting targets over $\mathcal{D}$ by $\mathcal{E}$ if there exist polynomials $g(n, 1/\epsilon)$ and $d(n, 1/\epsilon)$ such that for every $n \in \mathbb{N}$, $r_0 \in \mathcal{R}_n$, and $\epsilon > 0$, for any $\Delta \leq 1/d(n, 1/\epsilon)$, and every $\Delta$-drifting sequence $f_1, f_2, \ldots$ (with $f_i \in \mathcal{C}_n$ for all $i$), if $r_0, r_1, \ldots$ is generated by $\mathcal{E}$ such that $r_i = M(f_{i-1}, \mathcal{D}, \mathcal{E}, \epsilon, r_{i-1})$, then for all $\ell \geq g(n, 1/\epsilon)$, with probability at least $1 - \epsilon$, $\text{Perf}_{f_\ell}(r_\ell, \mathcal{D}) \geq 1 - \epsilon$. We refer to $d(n, 1/\epsilon)$ as the* drift polynomial.

As in the basic definition, we say that the class $\mathcal{C}$ is *evolvable with drifting targets over $\mathcal{D}$* if there exists a valid evolution algorithm $\mathcal{E} = (\mathcal{R}, \texttt{Neigh}, \mu, t, s)$ such that $\mathcal{C}$ is evolvable with drifting targets over $\mathcal{D}$ by $\mathcal{E}$. The drift polynomial specifies how much drift the algorithm can tolerate.

Our first main technical result, Theorem 8, relates the idea of monotonicity described above to drift resistance by showing that given a strictly beneficial neighborhood function for a class $\mathcal{C}$, distribution $\mathcal{D}$, and representation class $\mathcal{R}$, one can construct a mutation algorithm $\mathcal{E}$ such that $\mathcal{C}$ is evolvable with drifting targets over $\mathcal{D}$ by $\mathcal{E}$. The tolerance $t$ and sample size $s$ of $\mathcal{E}$ and the resulting generation polynomial $g$ and drift polynomial $d$ directly depend only on the benefit polynomial $b$ as described below. The proof is very similar to the proof of the first half of Lemma 5. Once again the key idea is to show that it is possible to set the tolerance such that with high probability, Bene is never empty and there is never a representation in Bene with performance too much worse than the guaranteed beneficial mutation. This implies that the mutation algorithm is guaranteed to choose a new representation with a significant increase in performance with respect to the previous target $f_{i-1}$ at each round $i$ with high probability. As long as $f_{i-1}$ and $f_i$ are sufficiently close, the chosen representation is also guaranteed to have good performance with respect to $f_i$.

**Theorem 8** *For any concept class $\mathcal{C}$, distribution $\mathcal{D}$, and representation class $\mathcal{R}$, if $\texttt{Neigh}$ is a strictly beneficial neighborhood function for $\mathcal{C}$, $\mathcal{D}$, and $\mathcal{R}$, then there exist valid functions $\mu$, $t$, and $s$ such that $\mathcal{C}$ is evolvable with drifting targets over $\mathcal{D}$ by $\mathcal{E} = (\mathcal{R}, \texttt{Neigh}, \mu, t, s)$. In particular, if $\texttt{Neigh}$ is strictly beneficial with benefit polynomial $b(n, 1/\epsilon)$, and $p(n, 1/\epsilon)$ is an arbitrary polynomial upper bound on the size of $\texttt{Neigh}(r, \epsilon)$, then $\mathcal{C}$ is evolvable with drifting targets over $\mathcal{D}$ with*

- *any distributions $\mu$ that satisfy $\mu(r, r', \epsilon) \geq 1/p(n, 1/\epsilon)$ for all $r \in \mathcal{R}_n$, $\epsilon$, and $r' \in \texttt{Neigh}(r, \epsilon)$,*
- *tolerance function $t(r, \epsilon) = 1/(2b(n, 1/\epsilon))$ for all $r \in \mathcal{R}_n$,*
- *any generation polynomial $g(n, 1/\epsilon) \geq 16b(n, 1/\epsilon)$,*
- *any sample size $s(n, 1/\epsilon) \geq 128(b(n, 1/\epsilon))^2 \ln\left(2p(n, 1/\epsilon)g(n, 1/\epsilon)/\epsilon\right)$, and*
- *any drift polynomial $d(n, 1/\epsilon) \geq 16b(n, 1/\epsilon)$, which allows drift $\Delta \leq 1/(16b(n, 1/\epsilon))$.*

In Sections 6 and 7, which can be read independent of Section 5, we appeal to this theorem in order to prove that some common concept classes are evolvable with drifting targets with relatively large values of $\Delta$. Using Lemma 5, we also obtain the following corollary.

**Corollary 9** *If a concept class $\mathcal{C}$ is strictly monotonically evolvable over $\mathcal{D}$, then $\mathcal{C}$ is evolvable with drifting targets over $\mathcal{D}$.*

## 5 Robustness Results

Feldman [9] proved that the original model of evolvability is equivalent to a restriction of the statistical query model of learning [15] known as learning by correlational statistical queries (CSQ) [5]. We extend Feldman's analysis to show that CSQ learning is also equivalent to both evolvability with drifting targets and quasi-monotonic evolvability, and so the notion of evolvability is robust to these changes in definition. We begin by briefly reviewing the CSQ model.

---
[3]Throughout the paper, we use the notation $\tilde{O}$ to suppress logarithmic factors.

## 5.1 Learning from Correlational Statistical Queries

The *statistical query* (SQ) model was introduced by Kearns [15] and has been widely studied due to its connections to learning with noise [1, 4]. Like the PAC model, the goal of an SQ learner is to produce a hypothesis $h$ that approximates the behavior of a target function $f$ with respect to a fixed but unknown distribution $\mathcal{D}$. Unlike the PAC model, the learner is not given direct access to labeled examples $\langle x, f(x) \rangle$, but is instead given access to a *statistical query oracle*. The learner submits queries of the form $(\psi, \tau)$ to the oracle, where $\psi : \mathcal{X} \times \{-1, 1\} \to [-1, 1]$ is a query function and $\tau \in [0, 1]$ is a tolerance parameter. The oracle responds to each query with any value $v$ such that $|E_{x \sim \mathcal{D}}[\psi(x, f(x))] - v| \leq \tau$. An algorithm is said to efficiently learn a class $\mathcal{C}$ in the SQ model if for all $n \in \mathbb{N}$, $\epsilon > 0$, and $f \in \mathcal{C}_n$, and every distribution $\mathcal{D}_n$ over $\mathcal{X}_n$, the algorithm, given access to $\epsilon$ and the SQ oracle for $f$ and $\mathcal{D}_n$, outputs a polynomially computable hypothesis $h$ in polynomial time such that $\text{err}(f, h) \leq \epsilon$. Furthermore it is required that each query $(\psi, \tau)$ made by the algorithm can be evaluated in polynomial time given access to $f$ and $\mathcal{D}_n$. It is known that any class efficiently learnable in the SQ model is efficiently learnable in the PAC model with label noise [15].

A query $(\psi, \tau)$ is called a *correlational statistical query* (CSQ) [5] if $\psi(x, f(x)) = \phi(x) f(x)$ for some function $\phi : \mathcal{X} \to [-1, 1]$. An algorithm $\mathcal{A}$ is said to efficiently learn a class $\mathcal{C}$ in the CSQ model if $\mathcal{A}$ efficiently learns $C$ in the SQ model using only correlational statistical queries.

It is useful to consider one additional type of query, the $\text{CSQ}_>$ query [9]. A $\text{CSQ}_>$ query is specified by a triple $(\phi, \theta, \tau)$, where $\phi : \mathcal{X} \to [-1, 1]$ is a query function, $\theta$ is a threshold, and $\tau \in [0, 1]$ is a tolerance parameter. When presented with such a query, a $\text{CSQ}_>$ oracle for target $f$ and distribution $\mathcal{D}$ returns 1 if $E_{x \sim \mathcal{D}}[\phi(x) f(x)] \geq \theta + \tau$, 0 if $E_{x \sim \mathcal{D}}[\phi(x) f(x)] \leq \theta - \tau$, and arbitrary value of either 1 or 0 otherwise. Feldman [9] showed that if there exists an algorithm for learning $\mathcal{C}$ over $\mathcal{D}$ that makes CSQs, then there exists an algorithm for learning $\mathcal{C}$ over $\mathcal{D}$ using $\text{CSQ}_>$s of the form $(\phi, \theta, \tau)$ where $\theta \geq \tau$ for all queries. Furthermore the number of queries made by this algorithm is at most $O(\log(1/\tau))$ times the number of queries made by the original CSQ algorithm.

## 5.2 Overview of the Reduction

The construction we present uses Feldman's simulation [9] repeatedly. Fix a concept class $\mathcal{C}$ and a distribution $\mathcal{D}$ such that $\mathcal{C}$ is learnable over $\mathcal{D}$ in the CSQ model. As mentioned above, this implies that there exists a $\text{CSQ}_>$ algorithm $\mathcal{A}$ for learning $\mathcal{C}$ over $\mathcal{D}$. Let $\mathcal{H}$ be the class of hypotheses from which the output of $\mathcal{A}$ is chosen. In the analysis that follows, we restrict our attention to the case in which $\mathcal{A}$ is deterministic. However, the extension of our analysis to randomized algorithms is straightforward using Feldman's ideas (see Lemma 4.7 in his paper [9]).

First, we present a high level outline of our reduction. Throughout this section we will use randomized Boolean functions. If $\psi : \mathcal{X} \to [-1, 1]$ is a real valued function, let $\Psi$ denote the randomized Boolean function such that for every $x$, $E[\Psi(x)] = \psi(x)$. It can be easily verified that for any function $\phi(x)$, $E_{x,\Psi}[\phi(x) \Psi(x)] = E_x[\phi(x) \psi(x)]$. For the rest of this section, we will abuse notation and simply write real-valued functions in place of the corresponding randomized Boolean functions.

Our representation is of the form $r = (1 - \epsilon/2) h + (\epsilon/2) \Phi$. Here $h$ is a hypothesis from $\mathcal{H}$ and $\Phi$ is function that encodes the state of the $\text{CSQ}_>$ algorithm that is being simulated. Feldman's simulation only uses the second part. Our simulation runs in perpetuity, restarting Feldman's simulation each time it has completed. Since the target functions are drifting over time, if $h$ has high performance with respect to the current target function, it will retain the performance for some time steps in the future, but not forever. During this time, Feldman's simulation on the $\Phi$ part produces a new hypothesis $h'$ which has high performance at the time this simulation is completed. At this time, we will transition to a representation $r' = (1 - \epsilon/2) h' + (\epsilon/2) \Phi$, where $\Phi$ is reset to start Feldman's simulation anew. Thus, although the target drifts, our simulation will continuously run Feldman's simulation to find a hypothesis that has a high performance with respect to the current target.

The rest of section 5 details the reduction. First, we show how a single run of $\mathcal{A}$ is simulated, which is essentially Feldman's reduction with minor modifications. Then we discuss how to restart this simulation once it has completed. This requires the addition of certain intermediate states to keep the reduction feasible in the evolution model. We also show that our reduction can be made quasi-monotonic. Finally, we show how all this can be done using a representation class that is independent of $\epsilon$, as is required. This last step is shown in the appendix.

## 5.3 Construction of the Evolutionary Algorithm

We describe the construction of our evolutionary algorithm $\mathcal{E}$. Let $\tau = \tau(n, 1/\epsilon)$ be a polynomial lower bound on the tolerance of the queries made by $\mathcal{A}$ when run with accuracy parameter $\epsilon/4$. Without loss of generality, we may assume all queries are made with this tolerance. Let $q = q(n, 1/\epsilon)$ be a polynomial upper bound on the number of queries made by $\mathcal{A}$, and assume that $\mathcal{A}$ makes exactly $q$ queries (if not, redundant queries can be added). Here, we allow our representation class to be dependent on $\epsilon$. However, this restriction may be removed (cf. Appendix A.7.1) In the remainder of this section we drop the subscripts $n$ and $\epsilon$, except where there is a possibility of confusion.

Following Feldman's notation, let $z$ denote a bit string of length $q$ which records the oracle responses to the queries made by $\mathcal{A}$; that is, the $i$th bit of $z$ is 1 if and only if the answer to the $i$th query is 1. Let $|z|$ denote the length of $z$, $z^i$ the prefix of $z$ of length $i$, and $z_i$ the $i$th bit of $z$. Since $\mathcal{A}$ is deterministic, the $i$th query made by $\mathcal{A}$ depends only on responses to the previous $i-1$ queries. We denote this query by $(\phi_{z^{i-1}}, \theta_{z^{i-1}}, \tau)$, with $\theta_{z^{i-1}} \geq \tau$, as discussed in Section 5.1. Let $h_z$ denote the final hypothesis output by $\mathcal{A}$ given query responses $z$. Since we have chosen to simulate $\mathcal{A}$ with accuracy parameter $\epsilon/4$, $h_z$ is guaranteed to satisfy $\text{Perf}_f(h_z, \mathcal{D}) \geq 1 - \epsilon/4$ for any function $f$ for which the query responses in $z$ are valid. Finally, let $\sigma$ denote the empty string.

For every $i \in \{1, \cdots, q\}$ and $z \in \{0,1\}^i$, we define $\Phi_z = (1/q) \sum_{j=1}^{i} \mathbb{I}(z_j = 1)\phi_{z^{j-1}}(x)$, where $\mathbb{I}$ is an indicator function that is 1 if its input is true and 0 otherwise. For any $h \in \mathcal{H}$, define $r_\epsilon[h, z] = (1 - \epsilon/2)h(x) + (\epsilon/2)\Phi_z(x)$. Recall that each of these real-valued functions can be treated as a randomized Boolean function as required by the evolution model. The performance of this function, which we use as our basic representation, is mainly determined by the performance of $h$, but by setting the tolerance parameter low enough, the $\Phi_z$ part can learn useful information about the (drifting) targets by simulating $\mathcal{A}$.

Let $\tilde{R}_\epsilon = \{r_\epsilon[h, z] \mid h \in \mathcal{H}, 0 \leq |z| \leq q-1\}$. The representations in $\tilde{R}_\epsilon$ will be used for simulating one round of $\mathcal{A}$. To reach a state where we can restart the simulation, we will need to add intermediate representations. These are defined below.

Let $tu(n, 1/\epsilon)$ be an upper bound on $\epsilon\theta_{z^i}/(8q)$ for all $i$ and $z^i$. (This will be a polynomial upper bound on all tolerances $t$ that we define below.) Assume for simplicity that $K = 2/tu(n, 1/\epsilon)$ is an integer. Let $w_0 = r_\epsilon[h, z]$, for some $h \in \mathcal{H}$ and $|z| = q$ ($w_0$ depends on $h$ and $z$, but to keep notation simple we will avoid subscripts). For $k = 1, \ldots, K$, define $w_k = (1 - k(tu(n, 1/\epsilon)/2))w_0$. Notice that $w_K = \mathbf{0}$, where $\mathbf{0}$ is a function that can be realized by a randomized function that ignores its input and predicts $+1$ or $-1$ randomly. Let $W_\epsilon = \{w_i \mid w_0 = r_\epsilon[h, z], h \in \mathcal{H}, |z| = q, i \in \{0, \ldots, K\}\}$. Finally define $\mathcal{R}_\epsilon = \tilde{R}_\epsilon \cup W_\epsilon$. For every representation $r_\epsilon[h, z] \in \tilde{R}_\epsilon$, we set

- $\texttt{Neigh}(r_\epsilon[h, z], \epsilon) = \{r_\epsilon[h, z], r_\epsilon[h, z0], r_\epsilon[h, z1]\}$,
- $\mu(r_\epsilon[h, z], r_\epsilon[h, z], \epsilon) = \eta$ and $\mu(r_\epsilon[h, z], r_\epsilon[h, z0], \epsilon) = \mu(r, r_\epsilon[h, z1], \epsilon) = (1-\eta)/2$,
- $t(r_\epsilon[h, z], \epsilon) = \epsilon\theta_{z^i}/(8q)$.

For the remaining representations $w_k \in W_\epsilon$, with $w_0 = r_\epsilon[h, z]$, we set

- $\texttt{Neigh}(w_K, \epsilon) = \{w_K, r_\epsilon[\mathbf{0}, \sigma]\}$ and $\texttt{Neigh}(w_k, \epsilon) = \{w_k, w_{k+1}, r_\epsilon[h_{z,\epsilon}, \sigma]\}$ for all $k < K$,
- $\mu(w_K, w_K) = \eta$ and $\mu(w_K, r_\epsilon[\mathbf{0}, \sigma]) = 1 - \eta$, and $\mu(w_k, w_k, \epsilon) = \eta^2$, $\mu(w_k, w_{k+1}, \epsilon) = \eta - \eta^2$, and $\mu(w_k, r_\epsilon[h_{z,\epsilon}, \sigma]) = 1 - \eta$ for all $k < K$,
- $t(w_k, \epsilon) = tu(n, 1/\epsilon)$.

Finally, let $\eta = \epsilon/(4q + 2K)$, $\tau' = \min\{(\epsilon\tau)/(2q), tu(n, 1/\epsilon)/8\}$, and $s = 1/(2(\tau')^2)\log((6q + 3K)/\epsilon)$. Let $\mathcal{E} = (\mathcal{R}_\epsilon, \texttt{Neigh}, \mu, t, s)$ with components defined as above. We show that $\mathcal{E}$ evolves $\mathcal{C}$ over $\mathcal{D}$ tolerating drift of $\Delta = (\epsilon\tau)/(4q + 2K + 2)$. This value of drift, while small, is an inverse polynomial in $n$ and $1/\epsilon$ as required. The point to note is that the evolutionary algorithm runs perpetually, while still maintaining high performance on any given round with high probability.

For any representation $r$, we denote by LPE the union of the low probability events that some estimates of performance are not within $\tau'$ of their true value, or that a mutation with relative probability less than $2\eta$ (either in $\texttt{Bene}$ or $\texttt{Neut}$) is selected over other mutations.

### 5.4 Simulating the CSQ$_>$ Algorithm for Drifting Targets

We now show that it is possible to simulate a CSQ$_>$ algorithm using an evolution algorithm $\mathcal{E}$ even when the target is drifting. However, if we simulate a query $(\phi, \theta, \tau)$ on round $i$, there is no guarantee that the answer to this query will remain valid in future rounds. The following lemma shows that by lowering the tolerance of the simulated query below the tolerance that is actually required by the CSQ$_>$ algorithm, we are able to generate a sequence of query answers that remain valid over many rounds. Specifically, it shows that if $v$ is a valid response for the query $(\phi, \theta, \tau/2)$ with respect to $f_i$, then $v$ is also a valid response for the query $(\phi, \theta, \tau)$ with respect to $f_j$ for any $j \in [i - \tau/(2\Delta), i + \tau/(2\Delta)]$.

**Lemma 10** *Let $f_1, f_2, \cdots$ be a $\Delta$-drifting sequence with respect to the distribution $\mathcal{D}$ over $\mathcal{X}$. For any tolerance $\tau$, any threshold $\theta$, any indices $i$ and $j$ such that $|i - j| \leq \tau/(2\Delta)$, and any function $\phi : \mathcal{X} \to [-1, 1]$, if $\mathrm{E}_{x \sim \mathcal{D}}[\phi(x)f_j(x)] \geq \theta + \tau$, then $\mathrm{E}_{x \sim \mathcal{D}}[\phi(x)f_i(x)] \geq \theta + \tau/2$. Similarly, if $\mathrm{E}_{x \sim \mathcal{D}}[\phi(x)f_j(x)] \leq \theta - \tau$, then $\mathrm{E}_{x \sim \mathcal{D}}[\phi(x)f_i(x)] \leq \theta - \tau/2$.*

We say that a string $z$ is *consistent* with a target function $f$, if for all $1 \leq i \leq |z|$, $z_i$ is a valid response to the query $(\phi_{z^{i-1}}, \theta_{z^{i-1}}, \tau)$, with respect to $f$. Suppose that the algorithm $\mathcal{E}$ starts with representation $r_0 = r_\epsilon[h, \sigma]$. (Recall that $\sigma$ denotes the empty string.) The following lemma shows that after $q$ time steps, with high probability it will reach a representation $r_\epsilon[h, z]$ where $|z| = q$ and $z$ is consistent with the target function $f_q$, implying that $z$ is a proper simulation of $\mathcal{A}$ on $f_q$.

**Lemma 11** *If $\Delta \leq \tau/(2q)$, then for any $\Delta$-drifting sequence $f_0, f_1, \ldots, f_q$, if $r_0, r_1, \ldots, r_q$ is the sequence of representations of $\mathcal{E}$ starting at $r_0 = r_\epsilon[h, \sigma]$, and if the LPE does not occur for $q$ rounds, then $r_q = r_\epsilon[h, z]$ where $|z| = q$ and $z$ is consistent with $f_q$.*

The proof uses the following ideas: If the LPE does not occur, there are no mutations of the form $r \to r$, so the length of $z$ increases by 1 every round, and also all estimates of performance are within $\tau'$ of their true value. When this is the case, and after observing that $r_\epsilon[h, z^i 0]$ is always neutral, it is possible to show that for any round $i$, (i) if $r_\epsilon[h, z^i 1]$ is beneficial, then 1 is a valid answer to the $i$th query with respect to $f_i$, (ii) if $r_\epsilon[h, z^i 1]$ is deleterious then 0 is a valid answer for the $i$th query with respect to $f_i$, and (c) if $r_\epsilon[h, z^i 1]$ is neutral, then both 0 and 1 are valid answers to the $i$th query. This implies that $z_{i+1}$ is always a valid answer to the $i$th query with respect to $f_i$, and by Lemma 10, with respect to $f_q$.

### 5.5 Restarting the Simulation

We now discuss how to restart Feldman's simulation once it completes. Suppose we are in a representation of the form $r_\epsilon[h, z]$, where $|z| = q$, and $z$ is consistent with the current target function $f$. Then if $h_z$ is the hypothesis output by $\mathcal{A}$ using query responses in $z$, we are guaranteed that (with high probability) $\mathtt{Perf}_f(h_z, \mathcal{D}) \geq 1 - \epsilon/4$. At this point, we would like the algorithm to choose a new representation $r_\epsilon[h_z, \sigma]$, where $\sigma$ is the empty string. The intuition behind this move is as follows. The performance of $r_\epsilon[h_z, \sigma]$ is guaranteed to be high (and to remain high for many generations) because much of the weight is on the $h_z$ term. Thus we can use the second term ($\Phi_\sigma$) to restart the learning process. After $q$ more time steps have passed, it may be the case that the performance of $h_z$ is no longer as high with respect to the new target, but the simulated algorithm will have already found a different hypothesis that does have high performance with respect to this new target.

There is one tricky aspect of this approach. In some circumstances, we may need to restart the simulation by moving from $r_\epsilon[h, z]$ to $r_\epsilon[h_z, \sigma]$ even though $z$ is *not* consistent with $f$. This situation can arise for two reasons. First, we might be near the beginning of the evolution process when $\mathcal{E}$ has not had enough generations to correctly determine the query responses (starting state may be $r_\epsilon[h, z_0]$ where $z_0$ has wrong answers). Second, there is some small probability of failure on any given round and we would like the evolutionary algorithm to recover from such failures smoothly. In either case, to handle the situation in which $h_z$ may have performance below zero (or very close), we will also allow $r_\epsilon[h, z]$ to mutate to $r_\epsilon[\mathbf{0}, \sigma]$.

The required changes from $r_\epsilon[h, z]$ to either $r_\epsilon[h_z, \sigma]$ or $r_\epsilon[\mathbf{0}, \sigma]$ described above may be deleterious. To handle this, we employ a technique of Feldman [9], where we first decrease the performance gradually (through neutral mutations) until these mutations are no longer deleterious. The representations defined in $W_\epsilon$ achieve this. The claim is that starting from any representation of the form $w_k$, we reach either $r_\epsilon[h_z, \sigma]$ or $r_\epsilon[\mathbf{0}, \sigma]$ in at most $K - k + 1$ steps, with high probability. Furthermore, since the probability of moving to $r_\epsilon[h_z, \sigma]$ is very high, this representation will be reached if it is ever a neutral mutation (i.e., the LPE does not happen). Thus, the performance always stays above the performance of $r_\epsilon[h_z, \sigma]$. Lemma 12 formalizes this claim.

**Lemma 12** *If $\Delta \leq tu(n, 1/\epsilon)/4$, then for any $\Delta$-drifting sequence $f_0, f_1, \ldots, f_q$, if $r_0, r_1, \ldots, r_q$ is the sequence of mutations of $\mathcal{E}$ starting at $r_0 = w_k$, then if the LPE does not happen at any time-step, there exists a $j \leq K - k + 1$ such that $r_j = r_\epsilon[h_{z, \epsilon}, \sigma]$ or $r_j = r_\epsilon[\mathbf{0}, \sigma]$. Furthermore, for all $1 \leq i < j$, $\mathtt{Perf}_{f_i}(r_i, \mathcal{D}) \geq \mathtt{Perf}_{f_i}(r_\epsilon[h_{z, \epsilon}, \sigma], \mathcal{D})$.*

### 5.6 Equivalence to Evolvability with Drifting Targets

Combining these results, we prove the equivalence between evolvability and evolvability with drifting targets starting from any representation in $\mathcal{R}_\epsilon$. The proof we give here uses the representation class $\mathcal{R}_\epsilon$ and therefore assumes that the value of $\epsilon$ is known. For the needed generalization to the case where $\mathcal{R} = \cup_\epsilon \mathcal{R}_\epsilon$, Feldman's backsliding trick [9] can be used to first reach a representation with zero performance, and then move to a representation in $\mathcal{R}_\epsilon$. Theorem 13 shows that every concept class that is learnable using CSQs (and thus every class that is evolvable) is evolvable with drifting targets.

**Theorem 13** *If $\mathcal{C}$ is evolvable over distribution $\mathcal{D}$, then $\mathcal{C}$ is evolvable with drifting targets over $\mathcal{D}$.*

**Proof:** Let $\mathcal{A}$ be a $\text{CSQ}_>$ algorithm for learning $\mathcal{C}$ over $\mathcal{D}$ with accuracy $\epsilon/4$. $\mathcal{A}$ makes $q = q(n, 1/\epsilon)$ queries of tolerance $\tau$ and outputs $h$ satisfying $\texttt{Perf}_f(h, \mathcal{D}) \geq 1 - \epsilon/4$. Let $\mathcal{E}$ be the evolutionary algorithm derived from $\mathcal{A}$ as described in Section 5.3. Recall that $K = 2/tu(n, 1/\epsilon)$, let $g = 2q + K + 1$. We show that starting from an arbitrary representation $r_0 \in \mathcal{R}_\epsilon$, with probability at least $1 - \epsilon$, $\texttt{Perf}_{f_g}(r_g, \mathcal{D}) \geq 1 - \epsilon$. This is sufficient to show that for all $\ell \geq g$, with probability at least $1 - \epsilon$, $\texttt{Perf}_{f_\ell}(r_\ell, \mathcal{D}) \geq 1 - \epsilon$, since we can consider the run of $\mathcal{E}$ starting from $r_{\ell-g}$.

With the setting of parameters as described in Section 5.3, with probability at least $1 - \epsilon$, the LPE does not occur for $g$ time steps, i.e., all estimates are within $\tau' = \min\{(\tau\epsilon)/(2q), tu(n, 1/\epsilon)/8\}$ of their true value and unlikely mutations (those with relative probabilities less that $2\eta$) are not chosen. Thus, we can apply the results of Lemmas 11 and 12. We assume that this is the case for the rest of the proof. When $\Delta = (\epsilon\tau)/(4q + 2K + 2)$, the assumption of Lemmas 11 and 12 hold and we can apply them.

First, we argue that starting from an arbitrary representation, in at most $q + K$ steps, we will have reached a representation of the form $r_\epsilon[h, \sigma]$, for some $h \in \mathcal{H}$. If the start representation is $r_\epsilon[h, z]$ for $|z| \leq q - 1$, then in at most $q - 1$ steps we reach a representation of the form $r_\epsilon[h, z']$ with $|z'| = q$, in which case by Lemma 12, the algorithm will transition to representation $r_\epsilon[h, \sigma]$ in at most $K + 1$ additional steps. Alternately, if the start representation is $w_k$ for $k \in \{0, \ldots, K\}$ as defined in Section 5.3, then by Lemma 12, we reach a representation of the form $r_\epsilon[h, \sigma]$ in at most $K + 1$ steps.

Let $m$ be the time step when $\mathcal{E}$ first reaches the representation of the form $r_\epsilon[h, \sigma]$. Then using Lemma 11, $r_{m+q} = r_\epsilon[h, z^*]$, where $z^*$ is consistent with $f_{m+q}$. Let $h^* = h_{z^*, \epsilon}$ be the hypothesis output by the simulated run of $\mathcal{A}$. Then $\texttt{Perf}_{f_{m+q}}(h^*, \mathcal{D}) \geq 1 - \epsilon/4$, and hence $\texttt{Perf}_{f_{m+q}}(r_\epsilon[h^*, \sigma], \mathcal{D}) \geq 1 - 3\epsilon/4$. For the value of $\Delta$ we are using, for all $i \leq g$, $\texttt{Perf}_{f_i}(r_\epsilon[h^*, \sigma], \mathcal{D}) \geq 1 - \epsilon$.

From such a representation, when all estimates of performance are within $\tau'$ of their true value and unlikely mutations (those with relative probability $\leq 2\eta$) do not occur, the performance will remain above $1 - \epsilon$. By Lemma 12, the algorithm will move from $r_{m+q} = r_\epsilon[h, z^*]$ to $r_\epsilon[h^*, \sigma]$ in at most $K + 1$ steps, and during these time steps for any time step $i$ it holds that $\texttt{Perf}_{f_i}(r_i, \mathcal{D}) \geq \texttt{Perf}_{f_i}(r_\epsilon[h^*, \sigma], \mathcal{D})$. Once $r_\epsilon[h^*, \sigma]$ is reached, for $q$ steps the representations will be of the form $r_\epsilon[h^*, z]$. For any such time step $i$, $\texttt{Perf}_{f_i}(r_i, \mathcal{D}) \geq \texttt{Perf}_{f_i}(r_\epsilon[h^*, \sigma], \mathcal{D})$. This is because if the answers in $z$ are correct (and they will be since the LPE does not happen at any time step), the term $\Phi_z$ is made up of only those functions $\phi_{z^{j-1}}$ for which $z^j = 1$, which are those for which $\phi_{z^{j-1}}$ has a correlation greater than $\theta_{z^{j-1}} - \tau \geq 0$ with the target $f_i$ (using Lemma 10). Since as observed above the performance of $r_\epsilon[h^*, \sigma]$ does not degrade below $1 - \epsilon$ in the time horizon we are interested in $\texttt{Perf}_{f_i}(r_i, \mathcal{D}) \geq \texttt{Perf}_{f_i}[r_\epsilon[h^*, \sigma]) \geq 1 - \epsilon$. ∎

### 5.7 Equivalence to Quasi-Monotonic Evolution

Finally, we show that all evolvable classes are also evolvable quasi-monotonically. In the proof of Theorem 13, we showed that for all $\ell \geq g = 2q + K + 1$, with high probability $\texttt{Perf}_{f_\ell}(r_\ell, \mathcal{D}) \geq 1 - \epsilon$, so quasi-monotonicity is satisfied trivially. Thus we only need to show quasi-monotonicity for the first $g$ steps. We will use the same construction as defined in Section 5.3, with modifications. However, this assumes that the representation knows $\epsilon$, since now the trick of having the performance slide back to zero would violate quasi-monotonicity. To make the representation class independent of $\epsilon$ a more complex construction is needed. Details can be found in the appendix.

**Theorem 14** *If $\mathcal{C}$ is evolvable over distribution $\mathcal{D}$, then $\mathcal{C}$ is quasi-monotonically evolvable over $\mathcal{D}$ with drifting targets.*

## 6 Evolving Hyperplanes with Drifting Targets

In this section, we present two alternative algorithms for evolving $n$-dimensional hyperplanes with drifting targets. The first algorithm, which generates the neighbors of a hyperplane by rotating it a small amount in one of $2(n-1)$ directions, tolerates drift on the order of $\epsilon/n$, but only over spherically symmetric distributions. The second algorithm, which generates the neighbors of a hyperplane by shifting single components of its normal vector, tolerates a smaller drift, but works when the distribution is an unknown product normal distribution. To our knowledge, these are the first positive results on evolving hyperplanes in the computational model of evolution.

Formally, let $\mathcal{C}_n$ be the class of all $n$-dimensional homogeneous linear separators.[4] For notational convenience, we reference each linear separator in $\mathcal{C}_n$ by the hyperplane's $n$-dimensional unit length normal vector $\mathbf{f} \in \mathbb{R}^n$. For every $\mathbf{f} \in \mathcal{C}_n$ and $\mathbf{x} \in \mathbb{R}^n$, we then have that $\mathbf{f}(\mathbf{x}) = 1$ if $\mathbf{f} \cdot \mathbf{x} \geq 0$, and $\mathbf{f}(\mathbf{x}) = -1$ otherwise. The evolution algorithms we consider in this section use a representation class $\mathcal{R}_n$ also consisting of $n$-dimensional unit vectors, where $\mathbf{r} \in \mathcal{R}_n$ is the normal vector of the hyperplane it represents.[5] Then $\mathcal{R} = \{\mathbf{r} \mid \|\mathbf{r}\|_2 = 1\}$. We describe the two algorithms in turn.

---
[4] A homogeneous linear separator is one that passes through the origin. [6]

## 6.1 An Evolution Algorithm Based on Rotations

For the rotation-based algorithm, we define the neighborhood function of $\mathbf{r} \in \mathcal{R}_n$ as follows. Let $\{\mathbf{u}^1 = \mathbf{r}, \mathbf{u}^2, \cdots, \mathbf{u}^n\}$ be an orthonormal basis for $\mathbb{R}^n$. This orthonormal basis can be chosen arbitrarily (and potentially randomly) as long as $\mathbf{u}^1 = \mathbf{r}$. Then

$$\texttt{Neigh}(\mathbf{r}, \epsilon) = \mathbf{r} \cup \left\{\mathbf{r}' \mid \mathbf{r}' = \cos\left(\epsilon/(\pi\sqrt{n})\right) \mathbf{r} \pm \sin\left(\epsilon/(\pi\sqrt{n})\right) \mathbf{u}^i \;,\; i \in \{2, \cdots, n\}\right\} \;. \quad (1)$$

In other words, each $\mathbf{r}' \in \texttt{Neigh}(\mathbf{r}, \epsilon)$ is obtained by rotating $\mathbf{r}$ by an angle of $\epsilon/(\pi\sqrt{n})$ in some direction. The size of this neighbor set is clearly $2n - 1$. We obtain the following theorem.

**Theorem 15** *Let $\mathcal{C}$ be the class of homogeneous linear separators, $\mathcal{R}$ be the class of homogeneous linear separators represented by unit length normal vectors, and $\mathcal{D}$ be an arbitrary spherically symmetric distribution. Define $\texttt{Neigh}$ as in Equation 1 and let $p$ be any polynomial satisfying $p(n, 1/\epsilon) \geq 2n-1$. Then $\mathcal{C}$ is evolvable with drifting targets over $\mathcal{D}$ by algorithm $\mathcal{A} = (\mathcal{R}, \texttt{Neigh}, \mu, t, s)$ with*

- *any distributions $\mu$ that satisfy $\mu(r, r', \epsilon) \geq 1/p(n, 1/\epsilon)$ for all $r \in \mathcal{R}_n$, $\epsilon$, and $r' \in \texttt{Neigh}(r, \epsilon)$,*
- *tolerance function $t(r, \epsilon) = \epsilon/(\pi^3 n)$ for all $r \in \mathcal{R}_n$,*
- *any generation polynomial $g(n, 1/\epsilon) \geq 8\pi^3 n/\epsilon$,*
- *a sample size $s(n, 1/\epsilon) = \tilde{O}(n^2/\epsilon^2)$, and*
- *any drift polynomial $d(n, 1/\epsilon) \geq 8\pi^3 n/\epsilon$, which allows drift $\Delta \leq \epsilon/(8\pi^3 n)$.*

To prove this, we need only to show that $\texttt{Neigh}$ is a strictly beneficial neighborhood function for $\mathcal{C}$, $\mathcal{D}$, and $\mathcal{R}$ with $b(n, 1/\epsilon) = \pi^3 n/(2\epsilon)$. The theorem then follows from Theorem 8. The analysis relies on the fact that under any spherically symmetric distribution $\mathcal{D}$ (for example, the uniform distribution over a sphere), $\texttt{err}_\mathcal{D}(\mathbf{u}, \mathbf{v}) = \arccos(\mathbf{u} \cdot \mathbf{v})/\pi$, where $\arccos(\mathbf{u} \cdot \mathbf{v})$ is the angle between $\mathbf{u}$ and $\mathbf{v}$ [6]. This allows us to reason about the performance of one function with respect to another by analyzing the dot product between their normal vectors.

## 6.2 A Component-Wise Evolution Algorithm

We now describe the alternate algorithm for evolving homogeneous linear separators. The guarantees we achieve are inferior to those described in the previous section. However, this algorithm applies when $\mathcal{D}$ is any unknown product normal distribution (with polynomial variance) over $\mathbb{R}^n$.

Let $r_i$ and $f_i$ denote the $i$th components of $\mathbf{r}$ and $\mathbf{f}$ respectively (not the values of the representation and ideal function at round $i$ as in previous sections). The alternate algorithm is based on the following observations. First, whenever there exists some $i$ for which $r_i$ and $f_i$ have different signs and aren't too close to 0, we can obtain a new representation with a non-trivial increase in performance by flipping the sign of $r_i$. Second, if there are no beneficial sign flips, if there is some $i$ for which $r_i$ is not too close to $f_i$, we can obtain a new representation with a significant increase in performance by adjusting $r_i$ a little and renormalizing. The amount we must adjust $r_i$ depends on the standard deviation of $\mathcal{D}$ in the $i$th dimension, so we must try many values when $\mathcal{D}$ is unknown. Finally, if the above conditions do not hold, then the performance of $r$ is already good enough.

Denote by $\{\mathbf{e}_i\}_{i=1}^n$ the basis of $\mathbb{R}^n$. Let $\sigma_1, \ldots, \sigma_n$ be the standard deviation of the distribution $\mathcal{D}$ in the $n$ dimensions. We assume that $1 \geq \sigma_i \geq (1/n)^k$ for some constant $k$ for all $i$, and that the algorithm is given access to the value of $k$, but not the particular values $\sigma_i$. We define the neighborhood function as $\texttt{Neigh}(\mathbf{r}, \epsilon) = N_\text{fl} \cup N_\text{sl}$, where $N_\text{fl} = \{\mathbf{r} - 2r_i\mathbf{e}_i \mid i = 1, \ldots, d\}$ is the set of representations obtained by flipping the sign of one component of $r$, and

$$N_\text{sl} = \left\{\frac{r \pm \frac{j\epsilon^2}{12n^k\sqrt{n}}\mathbf{e}_i}{\|r \pm \frac{j\epsilon^2}{12n^k\sqrt{n}}\mathbf{e}_i\|_2} \;\middle|\; i \in \{1, \cdots, d\}, j \in \{1, \cdots, 4n^k\}\right\}$$

is the set obtained by shifting each component by various amounts. We obtain the following.

---
[5]Technically we must assume that the representations $\mathbf{r} \in \mathcal{R}_n$ and input points $x \in \mathbb{R}^n$ are expressed to a fixed finite precision so that $\mathbf{r} \cdot \mathbf{x}$ is guaranteed to be computable in polynomial time, but for simplicity, in the analysis that follows, we treat both as simply vectors of real numbers.

**Theorem 16** *Let $\mathcal{C}$ be the class of homogeneous linear separators, and $\mathcal{R}$ be the class of homogeneous linear separators represented by unit length normal vectors, and $\mathcal{D}$ be a product normal distribution with (unknown) standard deviations $\sigma_1, \cdots, \sigma_n$ such that $1 \geq \sigma_i \geq (1/n)^k$ for all $i$ for a constant $k$. Define Neigh as above and let $p$ be any polynomial such that $p(n, 1/\epsilon) \geq 8n^{2k+1} + 2n$. Then $\mathcal{C}$ is evolvable with drifting targets over $\mathcal{D}$ by algorithm $\mathcal{A} = (\mathcal{R}, \text{Neigh}, \mu, t, s)$ with*

- *any distribution $\mu$ satisfying $\mu(r, r', \epsilon) \geq 1/p(n, 1/\epsilon)$ for all $r \in \mathcal{R}_n$ and $r' \in \text{Neigh}(r, \epsilon)$,*
- *tolerance function $t(r, \epsilon) = \epsilon^6/(288n)$,*
- *any generation polynomial $g(n, 1/\epsilon) \geq 2304n/\epsilon^6$,*
- *a sample size $s(n, 1/\epsilon) = \tilde{O}(n^2/\epsilon^{12})$, and*
- *any drift polynomial $d(n, 1/\epsilon) \geq 2304n/\epsilon^6$, which allows drift $\Delta \leq \epsilon^6/(2304n)$.*

The proof formalizes the set of observations described above, using them to show that Neigh is a strictly beneficial neighborhood function for $\mathcal{C}$, $\mathcal{D}$, and $\mathcal{R}$ with $b(n, 1/\epsilon) = 144n/\epsilon^6$. The theorem is then an immediate consequence of Theorem 8.

## 7 Evolving Conjunctions with Drifting Targets

We now show that conjunctions are evolvable with drifting targets over the uniform distribution with a drift of $O(\epsilon^2)$, independent of $n$. We begin by examining monotone conjunctions and prove that the neighborhood function defined by Valiant [19] is a strictly beneficial neighborhood function with $b(n, 1/\epsilon) = \epsilon^2/9$. Our proof uses techniques similar to those used in the simplified analysis of Valiant's algorithm presented by Diochnos and Turán [8]. By building on ideas from Jacobson [14], we extend this result to show that general conjunctions are evolvable with the same rate of drift.

### 7.1 Monotone Conjunctions

We represent monotone conjunctions using a representation class $\mathcal{R}$ where each $r \in \mathcal{R}$ is a subset of $\{1, \cdots, n\}$ such that $|r| \leq \log_2(3/\epsilon)$, representing the conjunction of the variables $x_j$ for all $j \in r$. We therefore allow the representation class to depend on $\epsilon$ in our analysis. This dependence is easy to remove (e.g., using Valiant's technique of allowing an initial phase in which the length of the representation decreases until it is below $\log_2(3/\epsilon)$ [19]), but simplifies presentation.

The neighborhood of a representation $r$ consists of the set of conjunctions that are formed by adding a variable to $r$, removing a variable from $r$, and swapping a variable in $r$ with a variable not in $r$, plus the representation $r$ itself. Formally, define the following three sets of conjunctions: $\mathcal{N}^+(r) = \{r \cup \{j\} | j \notin r\}$, $\mathcal{N}^-(r) = \{r \setminus \{j\} | j \in r\}$, and $\mathcal{N}^\pm(r) = \{r \setminus \{j\} \cup \{k\} | j \in S, k \notin S\}$. The neighborhood $\text{Neigh}(r, \epsilon)$ is then defined as follows. Let $q = \lceil \log_2(3/\epsilon) \rceil$. If $r$ is the empty set, then $\text{Neigh}(r, \epsilon) = \mathcal{N}^+(r) \cup r$. If $0 < |r| < q$, then $\text{Neigh}(r, \epsilon) = \mathcal{N}^+(r) \cup \mathcal{N}^-(r) \cup N^\pm(r) \cup r$. Finally, if $|r| = q$, then $\text{Neigh}(r, \epsilon) = \mathcal{N}^-(r) \cup \mathcal{N}^\pm(r) \cup r$. Note that the size of the neighborhood is bounded by $1 + n + n^2/4$ in the worst case; the combined size of the sets $\mathcal{N}^+(r)$ and $\mathcal{N}^-(r)$ is at most $n$, and the size of $N^\pm(r)$ is at most $n^2/4$. We obtain the following theorem.

**Theorem 17** *Let $\mathcal{C}$ be the class of monotone conjunctions, $\mathcal{R}$ be the class of monotone conjunctions of size at most $q = \lceil \log_2(3/\epsilon) \rceil$ represented as subsets of indices, and $\mathcal{D}$ be the uniform distribution. Define Neigh as above and let $p$ be any polynomial satisfying $p(n, 1/\epsilon) \geq 1 + n + n^2/4$. Then $\mathcal{C}$ is evolvable with drifting targets over $\mathcal{D}$ by algorithm $\mathcal{A} = (\mathcal{R}, \text{Neigh}, \mu, t, s)$ with*

- *any distributions $\mu$ that satisfy $\mu(r, r', \epsilon) \geq 1/p(n, 1/\epsilon)$ for all $r \in \mathcal{R}_n$, $\epsilon$, and $r' \in \text{Neigh}(r, \epsilon)$,*
- *tolerance function $t(r, \epsilon) = \epsilon^2/18$ for all $r \in \mathcal{R}_n$,*
- *any generation polynomial $g(n, 1/\epsilon) \geq 144/\epsilon^2$,*
- *a sample size $s(n, 1/\epsilon) = \tilde{O}(1/\epsilon^2)$, and*
- *any drift polynomial $d(n, 1/\epsilon) \geq 144/\epsilon^2$, which allows drift $\Delta \leq \epsilon^2/144$.*

To prove the theorem, we show that Neigh is a strictly beneficial target function with benefit polynomial $b(n, 1/\epsilon) = 9/\epsilon^2$ and once again appeal to Theorem 8. The proof is then essentially just a case-by-case analysis of the performance of the best $r' \in \text{Neigh}(r, \epsilon)$ for an exhaustive set of conditions on $r$ and $f$.

### 7.2 General Conjunctions

Jacobson [14] proposed an extension to the algorithm above that applies to general conjunctions. The key innovation in his algorithm is the addition of a fourth set $\mathcal{N}'(r)$ to the neighborhood or $r$, where each $r' \in \mathcal{N}'(r)$ is obtained by negating a subset of the literals in $r$. We show here that the drift rate of his construction can be analyzed in a similar way to the monotone case.

We represent general conjunctions using a representation class $\mathcal{R}$ where each $r \in \mathcal{R}$ is a subset of $\{1, \cdots, n\} \cup \{-1, \cdots, -n\}$ such that $|r| \leq \log_2(3/\epsilon)$. Here each $r$ represents the conjunction of literals $x_j$ for all positive $j \in r$ and negated literals $x_{-j}$ for all negative $j \in r$, and we restrict $\mathcal{R}$ so that it is never the case that both $j \in r$ and $-j \in r$. The dependence of this representation class on $\epsilon$ can be removed as before.

As before, the neighborhood of a representation $r$ includes the set of conjunctions that are formed by adding a variable to $r$, removing a variable from $r$, and swapping a variable in $r$ with a variable not in $r$, plus the representation $r$ itself. However, it now also includes a fourth set $\mathcal{N}'(r)$ of all conjunctions that can be obtained by negating a subset of the literals of $r$. The size of the set $\mathcal{N}'(r)$ is at most $2^q \leq 6/\epsilon$, so by a similar argument to the one above, the size of the neighborhood is bounded by $1 + 2n + n^2 + 6/\epsilon$. We obtain the following theorem.

**Theorem 18** *Let $\mathcal{C}$ be the class of conjunctions, $\mathcal{R}$ be the class of conjunctions of at most $q = \lceil \log_2(3/\epsilon) \rceil$ literals represented as above, and $\mathcal{D}$ be the uniform distribution. Define `Neigh` as above and let $p$ be any polynomial satisfying $p(n, 1/\epsilon) \geq 1 + 2n + n^2 + 6/\epsilon$. Then $\mathcal{C}$ is evolvable with drifting targets over $\mathcal{D}$ by $\mathcal{A} = (\mathcal{R}, \mathtt{Neigh}, \mu, t, s)$ with $\mu$, $t$, $g$, $s$, and $d$ as specified in Theorem 17.*

The proof uses many of the same ideas as the proof of Theorem 17. However, there are a few extra cases that need to be considered. First, if $f$ is a "long" conjunction, and $r$ contains at least one literal that is the negation of a literal in $f$, then we show that adding another literal to $r$ leads to a significant increase in performance. (If $r$ is already of maximum size, then the performance is already good enough.) Second, we show that if $f$ is "short" and $r$ contains at least one literal that is the negation of a literal in $f$, then there exists an $r' \in \mathcal{N}'(r)$ with significantly better performance. All other cases are identical to the monotone case.

## References

[1] J. Aslam and S. Decatur. Specification and simulation of statistical query algorithms for efficiency and noise tolerance. *Journal of Computer and System Sciences*, 56(2):191–208, 1998.

[2] P. L. Bartlett. Learning with a slowly changing distribution. In *COLT 5*, 1992.

[3] R. D. Barve and P. M. Long. On the complexity of learning from drifting distributions. *Information and Computation*, 138(2):101–123, 1997.

[4] A. Blum, A. Kalai, and H. Wasserman. Noise-tolerant learning, the parity problem, and the statistical query model. *Journal of the ACM*, 50(4):506–519, 2003.

[5] N. H. Bshouty and V. Feldman. On using extended statistical queries to avoid membership queries. *Journal of Machine Learning Research*, 2:359–395, 2002.

[6] S. Dasgupta. Coarse sample complexity bounds for active learning. In *NIPS 18*, 2005.

[7] S. Dasgupta, A. Kalai, and C. Monteleoni. Analysis of perceptron-based active learning. *Journal of Machine Learning Research*, pages 281–299, 2009.

[8] D. I. Diochnos and G. Turán. On evolvability: The swapping algorithm, product distributions, and covariance. In *Fifth Symposium on Stochastic Algorithms, Foundations and Applications*, 2009.

[9] V. Feldman. Evolvability from learning algorithms. In *Proceedings of ACM STOC 40*, 2008.

[10] V. Feldman. Robustness of evolvability. In *COLT 22*, 2009.

[11] V. Feldman. A complete characterization of statistical query learning with applications to evolvability. In *Proceedings of the IEEE Symposium on Foundation of Computer Science*, 2009.

[12] D. P. Helmbold and P. M. Long. Tracking drifting concepts by minimizing disagreements. *Machine Learning*, 14(1):27–46, 1994.

[13] W. Hoeffding. Probability inequalities for sums of bounded random variables. *Journal of the American Statistical Association*, 58(301):13–30.

[14] B. Jacobson. Personal communication, 2007.

[15] M. Kearns. Efficient noise-tolerant learning from statistical queries. *JACM*, 45(6):983–1006, 1998.

[16] A. Kuh, T. Petsche, and R. Rivest. Incrementally learning time-varying half-planes. In *NIPS 4*, 1991.

[17] L. Michael. Evolvability via the Fourier transform. *Theoretical Computer Science*, 2010. To appear.

[18] L. G. Valiant. A theory of the learnable. *Communications of the ACM*, 27(11):1134–1142, 1984.

[19] L. G. Valiant. Evolvability. *Journal of the ACM*, 56(1):1–21, 2009.

# A Additional Proofs

## A.1 Accuracy of the Empirical Performance

In order to prove Lemma 5 and Theorem 8, it is necessary to examine how close the empirical performance of a representation $r$ is to the representation's true performance. The following simple lemma shows that as long as the sample size $s(n, 1/\epsilon)$ is sufficiently large, the empirical performance of each representation will be close to the true performance with high probability.

**Lemma 19** *Consider any $r \in \mathcal{R}$ and $f \in \mathcal{C}$ and fix any $Z > 0$ and $\delta > 0$. Let $N$ be an upper bound on the size of the neighborhood $\texttt{Neigh}(r, 1/\epsilon)$. For each $r' \in \texttt{Neigh}(r, \epsilon)$, let $v(r')$ be the empirical performance of $r'$ with respect to $f$ on a sample of size $s \geq 2\ln(2N/\delta)/Z^2$. With probability $1 - \delta$, for all $r' \in \texttt{Neigh}(r, \epsilon)$, $|v(r') - \texttt{Perf}_f(r', \mathcal{D})| \leq Z$.*

**Proof:** Consider a particular $r' \in \texttt{Neigh}(r, \epsilon)$. By Hoeffding's inequality [13], for any $Z$, $\Pr(|v(r') - \texttt{Perf}_f(r', \mathcal{D})| \geq Z) \leq 2\exp(-sZ^2/2)$. The right hand side of this inequality is upper bounded by $\delta/N$ as long as $s \geq 2\ln(2N/\delta)/Z^2$, as we have assumed. The lemma then follows from a standard application of the union bound. ∎

## A.2 Proof of Lemma 5

Suppose that $\texttt{Neigh}$ is a strictly beneficial neighborhood function for $\mathcal{C}$, $\mathcal{D}$, and $\mathcal{R}$ with benefit polynomial $b(n, 1/\epsilon)$. To prove the first half of the lemma, we will construct an algorithm for strictly monotonically evolving $\mathcal{C}$ over $\mathcal{D}$. First, for any $r \in \mathcal{R}_n$ and $\epsilon > 0$, we set the tolerance at $t(r, \epsilon) = 1/(2b(n, 1/\epsilon))$. We then set $s(n, 1/\epsilon) = 128(b(n, 1/\epsilon))^2 \ln(2p(n, 1/\epsilon)/\delta)$ for a choice of $\delta$ that will be specified below. By Lemma 19, this guarantees that on a particular round $i$, with probability at least $1 - \delta$, for all $r \in \texttt{Neigh}(r_{i-1}, \epsilon)$, $|v(r) - \texttt{Perf}_f(r, \mathcal{D})| \leq 1/(8b(n, 1/\epsilon))$. For the remainder of this proof, we refer to this high probability event as the HPE.

For any fixed round $i$, consider first the case that $\texttt{Perf}_f(r_{i-1}, \mathcal{D}) \geq 1 - \epsilon/2$. Since $r_{i-1} \in \texttt{Neigh}(r_{i-1}, \epsilon)$, there is always at least one neutral mutation available and there could be a beneficial mutation, so $r_i$ will always be chosen from either $\texttt{Bene}$ or $\texttt{Neut}$. Consider an arbitrary $r_i$ chosen from $\texttt{Bene} \cup \texttt{Neut}$. If the HPE occurs, then

$$\begin{aligned}
(\texttt{Perf}_f(r_{i-1}, \mathcal{D}) - \texttt{Perf}_f(r_i, \mathcal{D})) &\leq \left(v(r_{i-1}) + \frac{1}{8b(n, 1/\epsilon)}\right) - \left(v(r_i) - \frac{1}{8b(n, 1/\epsilon)}\right) \\
&= (v(r_{i-1}) - v(r_i)) + \frac{1}{4b(n, 1/\epsilon)} \\
&\leq t(r, \epsilon) + \frac{1}{4b(n, 1/\epsilon)} = \frac{3}{4b(n, 1/\epsilon)} \leq \frac{3\epsilon}{8} \ .
\end{aligned}$$

The last inequality uses the fact that $1/b(n, 1/\epsilon) \leq \epsilon/2$. This must be the case to guarantee that an improvement of $1/b(n, 1/\epsilon)$ is possible when the performance is arbitrarily close to (but still less than) $1 - \epsilon/2$; otherwise, the definition of strictly beneficial neighborhood would not be satisfied. We then have

$$\texttt{Perf}_f(r_i, \mathcal{D}) \geq 1 - \frac{\epsilon}{2} - \frac{3\epsilon}{8} > 1 - \epsilon \ . \tag{2}$$

Now consider the case in which $\texttt{Perf}_f(r_{i-1}, \mathcal{D}) < 1 - \epsilon/2$. Since $\texttt{Neigh}$ is a strictly beneficial neighborhood function, it must be the case that there exists a representation $r \in \texttt{Neigh}(r_{i-1}, \epsilon)$ such that $\texttt{Perf}_f(r, \mathcal{D}) \geq \texttt{Perf}_f(r_{i-1}, \mathcal{D}) + 1/b(n, 1/\epsilon)$. Call this representation $r^*$. If the HPE occurs, then

$$\begin{aligned}
v(r^*) - v(r_{i-1}) &\geq \left(\texttt{Perf}_f(r^*, \mathcal{D}) - \frac{1}{8b(n, 1/\epsilon)}\right) - \left(\texttt{Perf}_f(r_{i-1}, \mathcal{D}) + \frac{1}{8b(n, 1/\epsilon)}\right) \\
&= (\texttt{Perf}_f(r^*, \mathcal{D}) - \texttt{Perf}_f(r_{i-1}, \mathcal{D})) - \frac{1}{4b(n, 1/\epsilon)} \geq \frac{3}{4b(n, 1/\epsilon)} > t(r, \epsilon) \ ,
\end{aligned}$$

and so $r^* \in \texttt{Bene}$. Since the set $\texttt{Bene}$ is non-empty, a representation in this set will be chosen for $r_i$. Consider an arbitrary $r_i$ chosen from $\texttt{Bene}$. If the HPE occurs, then

$$\begin{aligned}
(\texttt{Perf}_f(r_i, \mathcal{D}) - \texttt{Perf}_f(r_{i-1}, \mathcal{D})) &\geq \left(v(r) - \frac{1}{8b(n, 1/\epsilon)}\right) - \left(v(r_{i-1}) + \frac{1}{8b(n, 1/\epsilon)}\right) \\
&= (v(r) - v(r_{i-1})) - \frac{1}{4b(n, 1/\epsilon)} \\
&\geq t(r, \epsilon) - \frac{1}{4b(n, 1/\epsilon)} = \frac{1}{4b(n, 1/\epsilon)} \ . \tag{3}
\end{aligned}$$

Now, let $g(n, 1/\epsilon) = 8b(n, 1/\epsilon)$. Setting the parameter $\delta = \epsilon/g(n, 1/\epsilon)$ above and applying the union bound again, we have that with probability at least $1 - \epsilon$, the HPE occurs at all round $i \in \{1, 2, \cdots, g(n, 1/\epsilon)\}$. Suppose this is the case. From the argument leading up to Equation 3, we know that the performance of the current representation is monotonically increasing as long as the performance is less than $1 - \epsilon/2$, and furthermore increases by at least $1/(4b(n, 1/\epsilon))$ on each around. It remains to show that the algorithm evolves $\mathcal{C}$, that is, that $\text{Perf}_f(r_{g(n,1/\epsilon)}, \mathcal{D}) \geq 1 - \epsilon$.

From the monotonic improvement when the performance is less than $1 - \epsilon/2$ and the argument leading up to Equation 2, it is clear that if the performance ever reaches $1 - \epsilon/2$, it will not fall below $1 - \epsilon$ again before round $g(n, 1/\epsilon)$. It is easy to see that the performance reaches $1 - \epsilon/2$ at some point during these $g(n, 1/\epsilon)$ rounds. In the worst case, the performance starts at $-1$. It is guaranteed to increase by at least $1/(4b(n, 1/\epsilon))$ on each round. Thus it must reach $1 - \epsilon/2$ in no more than $g(n, 1/\epsilon) = 8b(n, 1/\epsilon)$ rounds.

To prove the second half of the lemma, note that the definition of strictly monotonic evolvability requires that for any initial representation $r_0 \in \mathcal{R}$, with high probability either $\text{Perf}_f(r_0, \mathcal{D}) \geq 1 - \epsilon/2$ or $\text{Perf}_f(r_1, \mathcal{D}) \geq \text{Perf}_f(r_0, \mathcal{D}) + 1/m(n, 2/\epsilon)$. This implies that if $\text{Perf}_f(r_0, \mathcal{D}) < 1 - \epsilon/2$ there must exist an $r_1 \in \text{Neigh}(r_1, \epsilon)$ such that $\text{Perf}_f(r_1, \mathcal{D}) \geq \text{Perf}_f(r_0, \mathcal{D}) + 1/m(n, 2/\epsilon)$. ∎

### A.3 Proof of Theorem 8

Suppose that Neigh is a strictly beneficial neighborhood function for $\mathcal{C}$, $\mathcal{D}$, and $\mathcal{R}$ with benefit polynomial $b(n, 1/\epsilon)$. For any $r \in \mathcal{R}_n$ and $\epsilon > 0$, we set the tolerance at $t(r, \epsilon) = 1/(2b(n, 1/\epsilon))$. We then set $s(n, 1/\epsilon) = 128(b(n, 1/\epsilon))^2 \ln(2p(n, 1/\epsilon)/\delta)$ for a choice of $\delta$ that will be specified below. This guarantees that on a particular round $i$, with probability at least $1 - \delta$, for all $r \in \text{Neigh}(r_{i-1}, \epsilon)$, $|v(r) - \text{Perf}_{f_{i-1}}(r, \mathcal{D})| \leq 1/(8b(n, 1/\epsilon))$. (See Lemma 19 in Appendix A.1 for details.) For the remainder of this proof, we refer to this high probability event as the HPE.

Fix an $i$. Suppose $\Delta \leq 1/(16b(n, 1/\epsilon))$. If $f_1, f_2, \cdots$ is a $\Delta$-drifting sequence, then for any $r \in \mathcal{R}$,

$$\begin{aligned}
\left|\text{Perf}_{f_{i-1}}(r, \mathcal{D}) - \text{Perf}_{f_i}(r, \mathcal{D})\right| &\leq \mathbb{E}_{x \sim \mathcal{D}}\big[|f_{i-1}(x) - f_i(x)| \cdot |r(x)|\big] \leq 2\text{err}_\mathcal{D}(f_{i-1}, f_i) \\
&\leq 2\Delta \leq 1/(8b(n, 1/\epsilon)) .
\end{aligned} \qquad (4)$$

Consider the case that $\text{Perf}_{f_{i-1}}(r_{i-1}, \mathcal{D}) \geq 1 - \epsilon/2$. Since $r_{i-1} \in \text{Neigh}(r_{i-1}, \epsilon)$, there is at least one neutral mutation available and there could be a beneficial mutation, so $r_i$ will be chosen from either Bene or Neut. Consider an arbitrary $r_i$ chosen from Bene $\cup$ Neut. If the HPE occurs, then

$$\begin{aligned}
\text{Perf}_{f_{i-1}}(r_{i-1}, \mathcal{D}) - \text{Perf}_{f_{i-1}}(r_i, \mathcal{D}) &\leq \left(v(r_{i-1}) + \frac{1}{8b(n, 1/\epsilon)}\right) - \left(v(r_i) - \frac{1}{8b(n, 1/\epsilon)}\right) \\
&\leq t(r, \epsilon) + \frac{1}{4b(n, 1/\epsilon)} = \frac{3}{4b(n, 1/\epsilon)} .
\end{aligned}$$

Then from Equation 4 and the assumption that $\Delta \leq 1/(16b(n, 1/\epsilon))$,

$$\begin{aligned}
\text{Perf}_{f_i}(r_i, \mathcal{D}) &\geq \text{Perf}_{f_{i-1}}(r_i, \mathcal{D}) - \frac{1}{8b(n, 1/\epsilon)} \\
&\geq \left(\text{Perf}_{f_{i-1}}(r_{i-1}, \mathcal{D}) - \frac{3}{4b(n, 1/\epsilon)}\right) - \frac{1}{8b(n, 1/\epsilon)} \geq \left(1 - \frac{\epsilon}{2}\right) - \frac{7\epsilon}{16} > 1 - \epsilon . \qquad (5)
\end{aligned}$$

The last line uses the fact that $1/b(n, 1/\epsilon) \leq \epsilon/2$. This must be the case to guarantee that an improvement of $1/b(n, 1/\epsilon)$ is possible when the performance is arbitrarily close to (but still less than) $1 - \epsilon/2$; otherwise, the definition of strictly beneficial neighborhood would not be satisfied.

Now consider the case in which $\text{Perf}_{f_{i-1}}(r_{i-1}, \mathcal{D}) < 1 - \epsilon/2$. Since Neigh is a strictly beneficial neighborhood function, it must be the case that there exists a representation $r \in \text{Neigh}(r_{i-1}, \epsilon)$ such that $\text{Perf}_{f_{i-1}}(r, \mathcal{D}) \geq \text{Perf}_{f_{i-1}}(r_{i-1}, \mathcal{D}) + 1/b(n, 1/\epsilon)$. Call this $r^*$. If the HPE occurs, then

$$\begin{aligned}
v(r^*) - v(r_{i-1}) &\geq \left(\text{Perf}_{f_{i-1}}(r^*, \mathcal{D}) - \frac{1}{8b(n, 1/\epsilon)}\right) - \left(\text{Perf}_{f_{i-1}}(r_{i-1}, \mathcal{D}) + \frac{1}{8b(n, 1/\epsilon)}\right) \\
&= \left(\text{Perf}_{f_{i-1}}(r^*, \mathcal{D}) - \text{Perf}_{f_{i-1}}(r_{i-1}, \mathcal{D})\right) - \frac{1}{4b(n, 1/\epsilon)} \geq \frac{3}{4b(n, 1/\epsilon)} > t(r, \epsilon) ,
\end{aligned}$$

and so $r^* \in$ Bene. Since the set Bene is non-empty, a representation in this set will be chosen for $r_i$. Consider an arbitrary $r_i$ chosen from Bene. If the HPE occurs, then

$$\begin{aligned}
\text{Perf}_{f_{i-1}}(r_i, \mathcal{D}) - \text{Perf}_{f_{i-1}}(r_{i-1}, \mathcal{D}) &\geq (v(r) - 1/(8b(n, 1/\epsilon))) - (v(r_{i-1}) + 1/(8b(n, 1/\epsilon))) \\
&\geq t(r, \epsilon) - 1/(4b(n, 1/\epsilon)) = 1/(4b(n, 1/\epsilon)) .
\end{aligned}$$

Then from Equation 4,
$$\texttt{Perf}_{f_i}(r, \mathcal{D}) - \texttt{Perf}_{f_{i-1}}(r_{i-1}, \mathcal{D}) \geq \texttt{Perf}_{f_{i-1}}(r, \mathcal{D}) - \texttt{Perf}_{f_{i-1}}(r_{i-1}, \mathcal{D}) - 2\Delta$$
$$\geq 1/(4b(n, 1/\epsilon)) - 1/(8b(n, 1/\epsilon)) = 1/(8b(n, 1/\epsilon)) . \quad (6)$$

Now, let $g(n, 1/\epsilon) = 16b(n, 1/\epsilon)$ and consider any round $\ell \geq g(n, 1/\epsilon)$. Setting the parameter $\delta = \epsilon/g(n, 1/\epsilon)$ above and applying the union bound again, we have that with probability at least $1 - \epsilon$, the HPE occurs at all rounds $i \in \{\ell - g(n, 1/\epsilon), \cdots, \ell - 1\}$. Suppose this is the case.

From the argument leading up to Equation 6, we know that the performance of the current representation with respect to the current target is monotonically increasing as long as the performance is less than $1 - \epsilon/2$, and increases by at least $1/(8b(n, 1/\epsilon))$ on each round. Combining this with the argument leading up to Equation 5, it is clear that if the performance ever reaches $1 - \epsilon/2$ during this period of time, it will never again fall below $1 - \epsilon$ before round $\ell$. It remains to show that the performance reaches $1 - \epsilon/2$ at some point during these $g(n, 1/\epsilon)$ rounds. This is also easy to see. In the worst case, the performance starts at $-1$. It is guaranteed to increase by at least $1/(8b(n, 1/\epsilon))$ on each round, so it must reach $1 - \epsilon/2$ in no more than $g(n, 1/\epsilon) = 16b(n, 1/\epsilon)$ rounds.

This shows that for any $\ell \geq g(n, 1/\epsilon)$, with probability at least $1 - \epsilon$, $\texttt{Perf}_{f_\ell}(r_\ell, \mathcal{D}) \geq 1 - \epsilon$ and so $\mathcal{C}$ is evolvable with drifting targets. ∎

### A.4 Proof of Lemma 10

Assume that $i < j$. The proof for the case in which $i > j$ is nearly identical, and the result is trivial if $i = j$. For any $\tau$ and any function $\phi : \mathcal{X} \to [-1, 1]$,

$$|\mathrm{E}[\phi(x)f_i(x)] - \mathrm{E}[\phi(x)f_j(x)]| = |\mathrm{E}[\phi(x)(f_i(x) - f_j(x))]| = \left|\mathrm{E}\left[\phi(x)\sum_{k=1}^{j-i}(f_{i+k-1}(x) - f_{i+k}(x))\right]\right|$$

$$\leq \mathrm{E}\left[|\phi(x)|\sum_{k=1}^{j-i}|f_{i+k-1}(x) - f_{i+k}(x)|\right]$$

$$\leq \sum_{k=1}^{j-i}\mathrm{E}[|f_{i+k-1}(x) - f_{i+k}(x)|] \leq (j - i)\Delta \leq \frac{\tau}{2},$$

where all expectations are taken with respect to $x \sim \mathcal{D}$. Therefore if $\mathrm{E}_{x \sim \mathcal{D}}[\phi(x)f_j(x)] \geq \theta + \tau$, then
$$\mathrm{E}_{x \sim \mathcal{D}}[\phi(x)f_i(x)] \geq \mathrm{E}_{x \sim \mathcal{D}}[\phi(x)f_j(x)] - \frac{\tau}{2} \geq \theta + \tau - \frac{\tau}{2} = \theta + \frac{\tau}{2}.$$

Similarly, if $\mathrm{E}_{x \sim \mathcal{D}}[\phi(x)f_j(x)] \leq \theta - \tau$, then
$$\mathrm{E}_{x \sim \mathcal{D}}[\phi(x)f_i(x)] \leq \mathrm{E}_{x \sim \mathcal{D}}[\phi(x)f_j(x)] + \frac{\tau}{2} \leq \theta - \tau + \frac{\tau}{2} = \theta - \frac{\tau}{2}.$$
∎

### A.5 Proof of Lemma 11

Under the assumption that the LPE does not occur at any time step, after $q$ time steps if $r_q = r_\epsilon[h, z]$, then $|z| = q$, since we add one bit at each step. Let $r_i = r_\epsilon[h, z^i]$. We consider the possible mutations of $r_i$. Observe that $r_\epsilon[h, z^i 0]$ is always neutral for all $i$. The cases we need to consider are (a) $r_\epsilon[h, z^i 1]$ is beneficial, and therefore chosen as the next representation implying that $z_{i+1} = 1$, (b) $r_\epsilon[h, z^i 1]$ is deleterious, and therefore $r_\epsilon[h, z^i 0]$ is chosen as the next representation, implying that $z_{i+1} = 0$, and (c) $r_\epsilon[h, z^i 1]$ is neutral, which implies that either $r_\epsilon[h, z^i 1]$ or $r_\epsilon[h, z^i 0]$ can be chosen as the next representation, implying that $z_{i+1} = 0$ or $1$.

Suppose we are in case (a), then we show that 1 is a valid answer to the query $(\phi_{z^i, \epsilon}, \theta_{z^i, \epsilon}, \tau/2)$ with respect to $f_i$. Consider,

$$t\left(r_i, \frac{1}{\epsilon}\right) \leq v(r_\epsilon[h, z^i 1]) - v(r_i) \leq \texttt{Perf}_{f_i}(r_\epsilon[h, z^i 1], \mathcal{D}) - \texttt{Perf}_{f_i}(r_i, \mathcal{D}) + 2\tau' = \frac{\epsilon}{2q}\mathrm{E}[\phi_{z^i, \epsilon} \cdot f_i] + 2\tau' .$$

Re-arranging the terms, we get:
$$\mathrm{E}[\phi_{z^i, \epsilon} \cdot f_i] \geq \frac{2q}{\epsilon}\left(t\left(r_i, \frac{1}{\epsilon}\right) - 2\tau'\right) \geq \theta_{z^i, \epsilon} - \frac{\tau}{2} .$$

Similarly one can show in case (b), that $\mathrm{E}[\phi_{z^i, \epsilon} \cdot f] \leq \theta_{z^i, \epsilon} - \tau/2$ and hence 0 is a valid answer to the query $(\phi_{z^i, \epsilon}, \theta_{z^i, \epsilon}, \tau/2)$ and in case (c), that $\theta_{z^i, \epsilon} - \tau/2 \leq \mathrm{E}[\phi_{z^i, \epsilon} \cdot f] \leq \theta_{z^i, \epsilon} + \tau/2$, and hence both 0 and 1 are valid answers.

By Lemma 10, if $z_{i+1}$ is a valid answer to the query $(\phi_{z^i, \epsilon}, \theta_{z^i, \epsilon}, \tau/2)$, with respect to $f_i$ it is a valid answer to the $(\phi_{z^i, \epsilon}, \theta_{z^i, \epsilon}, \tau)$ with respect to $f_q$ (since $\Delta \leq \tau/(2q)$). Thus $z$ is consistent with $f_q$. ∎

## A.6 Proof of Lemma 12

Let $\tau' = tu(n, 1/\epsilon)/8$. Assuming that the LPE does not occur at any time step, $w_{j+1}$ is always a neutral mutation for $w_j$, and mutations of the form $w_j \to w_j$ will not occur. Also $r_\epsilon[h_{z,\epsilon}, \sigma]$ will always be chosen if it is a neutral mutation. Then in $K - k$ rounds we will reach $w_K$ (if we had not already gone to $r_\epsilon[h_{z,\epsilon}, \sigma]$) and hence on the next round we will move to $r_\epsilon[\mathbf{0}, \sigma]$. This implies that the number of steps is at most $K - k + 1$.

Now, suppose that if at some stage $\text{Perf}_{f_i}(r_i, \mathcal{D}) < \text{Perf}_{f_i}(r_\epsilon[h_{z,\epsilon}, \sigma], \mathcal{D})$. Then

$$\text{Perf}_{f_{i-1}}(r_\epsilon[h_{z,\epsilon}, \sigma], \mathcal{D}) - \text{Perf}_{f_{i-1}}(r_{i-1}, \mathcal{D})$$

$$\geq \text{Perf}_{f_i}(r_\epsilon[h_{z,\epsilon}, \sigma], \mathcal{D}) - \text{Perf}_{f_{i-1}}(r_i, \mathcal{D}) - \frac{tu(n, \frac{1}{\epsilon})}{2} - \Delta \geq -\frac{tu(n, \frac{1}{\epsilon})}{2} - \Delta,$$

and so $r_\epsilon[h_{z,\epsilon}, \sigma]$ is a neutral mutation for $r_{i-1}$. By the assumption above, $r_i = r_\epsilon[h_{z,\epsilon}, \sigma]$, proving the lemma. ∎

## A.7 Proof Sketch for Theorem 14

We apply pieces of analysis of Theorem 13 here. We omit some details since the arguments are very similar; in fact, the argument that this algorithm is resistant to drift is nearly identical. To start, we let the representation class be $\mathcal{R}_\epsilon$ which depends on $\epsilon$. Here, backsliding is not allowed since it degrades performance arbitrarily. We discuss how to encode all values of $\epsilon$ in the same representation class in the Section A.7.1 below.

We will use the same construction as defined in Section 5.3, with only a small modification. For the representations in $W_\epsilon$, say $w_0 = r_\epsilon[h, z]$ with $|z| = q$, and in the neighborhood of $w_k$ we will also add $r_\epsilon[h, \sigma]$ in addition to the existing $r_\epsilon[h_z, \sigma], r_\epsilon[\mathbf{0}, \sigma]$ and $w_{k+1}$. Thus, even if $h_{z^*}$ has poor performance, we can ensure that the performance goes more than $\epsilon$ lower than the starting state. Formally,

- $\text{Neigh}'(w_K, \epsilon) = \{w_K, r_\epsilon[\mathbf{0}, \sigma]\}$ and $\text{Neigh}'(w_k, \epsilon) = \{w_k, w_{k+1}, r_\epsilon[h_z, \sigma], r_\epsilon[h, \sigma]\}$ for $k < K$,
- $\mu'(w_K, w_K) = \eta$ and $\mu'(w_K, r_\epsilon[\mathbf{0}, \sigma]) = 1 - \eta$, and $\mu'(w_k, w_k) = \eta^2$, $\mu'(w_k, w_{k+1}) = \mu'(w_k, r_\epsilon[h_z, \sigma]) = (\eta - \eta^2)/2$, and $\mu'(w_k, r_\epsilon[h, \sigma]) = 1 - \eta$ for $k < K$, and
- $t'(w_k, \epsilon) = tu(n, 1/\epsilon)$ for all $k$.

Let $\eta = \epsilon/(4q + 2K)$, $\tau' = \min\{(\epsilon\tau)/(2q), tu(n, 1/\epsilon)/8\}$, and $s = 1/(2(\tau')^2)\log((6q + 3K)/\epsilon)$, as defined earlier in section 5.3. Let $\text{Neigh}' = \text{Neigh}$, $\mu' = \mu$ and $t' = t$ for the representations in $\tilde{R}_\epsilon$. We will show that $\mathcal{E} = (\mathcal{R}_\epsilon, \text{Neigh}', \mu', t', s)$ evolves $\mathcal{C}$ quasi-monotonically.

The intuition of the proof is as follows. Any two representations $r_\epsilon[h, z]$ and $r_\epsilon[h, z']$ are within performance $\epsilon$ of each other (by definition). Using a similar argument as that for Lemma 12, one can show that while we start decreasing performance from $w_0 = r_\epsilon[h, z]$, the performance never dips below the performance of $r_\epsilon[h, \sigma]$ (and since this has the highest probability, this will be chosen whenever it is neutral). If $r_\epsilon[h_z, \sigma]$ is chosen earlier, it will be because its performance was higher than that of $r_\epsilon[h, \sigma]$ and quasi-monotonicity is maintained. Just as in Theorem 13, one can show that in at most $2q + K + 1$ steps, a representation with performance $1 - \epsilon$ is reached.

For our setting of parameters, for $g$ time steps LPE does not occur. And we will assume that this is the case. Recall that $\Delta = (\epsilon\tau)/(4q + 2K + 2)$.

There are two distinct types of starting representations: (i) $r_\epsilon[h, z]$ with $|z| < q$, or (ii) $w_k$ for some $k$ where $w_0 = r_\epsilon[h, z]$ with $|z| = q$. Suppose first that the starting representation is $r_\epsilon[h, z]$. Since LPE events don't occur, we will reach $r_\epsilon[h, z^*]$ with $|z^*| = q$ in $q - |z|$ steps. Note that for all $z'$ for any $f$, $|\text{Perf}_f(r_\epsilon[h, z'], \mathcal{D}) - \text{Perf}_f(r_\epsilon[h, z], \mathcal{D})| \leq \epsilon$. So during this phase quasi-monotonicity is maintained.

Consider the case in which the starting representation is instead $w_k$ for some $k$, with $w_0 = r_\epsilon[h, z^*]$ and $|z^*| = q$, or the case in which we reach such a representation $w_k$ after starting at $r_\epsilon[h, z]$. The algorithm then transitions to either representation $r_\epsilon[h, \sigma]$ or representation $r_\epsilon[h_{z^*}, \sigma]$. Furthermore, the transition to $r_\epsilon[h_{z^*}, \sigma]$ happens only if $\text{Perf}_f(h_{z^*}, \mathcal{D}) \geq \text{Perf}_f(h, \mathcal{D})$. This happens in at most $K$ steps (using argument similar to that of Lemma 12) and during this time the performance never goes below that of $r_\epsilon[h, \sigma]$, and so never goes more than $\epsilon/2$ below that of the starting representation.

Let $h'$ be either $h$ or $h_{z^*}$ depending on which was chosen as described in the above paragraph. Since $\Delta$ is low, the performance of $h'$ will never go significantly below that of $h$ (even if $h' = h_{z^*}$) for the next $g$ steps, hence it is sufficient to prove then that the performance will not drop significantly below that of $r_\epsilon[h', \sigma]$. From $r_\epsilon[h', \sigma]$ in at most $q$ steps we reach a representation $r_\epsilon[h', z']$ where $z'$ is consistent with $f$. During this time the performance never goes more than $\epsilon/2$ below that of $\text{Perf}(r_\epsilon[h', \sigma], \mathcal{D})$. From $r_\epsilon[h', z']$ we reach a representation with performance greater than $1 - \epsilon$ in one step, or $r_\epsilon[h', z']$ already has performance at least $1 - \epsilon$. Thus, the evolution is quasi-monotonic. ∎

### A.7.1 Removing the Need to Know $\epsilon$

In Section A.7, we showed that any CSQ algorithm can be converted into an evolutionary algorithm that is drift-resistant and quasi-monotonic, provided we are allowed to fix $\epsilon$ and encode its value in the representation. Here, we describe in some detail how a representation class that simultaneously encodes all values of $\epsilon$ can be constructed. Note that the definition of evolvability allows the neighborhood to depend on $\epsilon$, but not the starting representation.

We assume that the parameter $\epsilon$ provided to the algorithm is a power of 2. If this were not the case we could simply run the algorithm with $\epsilon'$, setting $\epsilon' = 2^{\lfloor \log \epsilon \rfloor}$. The performance guarantees would only be better since $\epsilon' \le \epsilon$. Furthermore, since $\epsilon' \ge \epsilon/2$, the running time would not be affected, except up to a constant factor. The representations will encode values of $\epsilon$ ranging over the set $S_\epsilon = \{1/2, 1/4, \ldots, 2^{-n}\}$. It is not necessary to consider values of $\epsilon$ smaller than this, since this would allow the algorithm to take time exponential in $n$, and hence an exhaustive search over all functions of polynomial-sized representations would be feasible in just one round of evolution. For the rest of this section, assume that $\epsilon$ can only take values from this set.

Recall the notation used in Section 5. In particular, $\mathcal{A}$ is a $CSQ_>$ algorithm that takes parameter $\epsilon$, makes $q = q(n, 1/\epsilon)$ queries of tolerance $\tau(n, 1/\epsilon)$, returns a hypothesis $h$ with $\texttt{Perf}_f(h, \mathcal{D}) \ge 1 - \epsilon$. Similar to the definitions in 5.3, let $\Phi_{z,\epsilon} = (1/q) \sum_{i=1}^{q} \mathbb{I}(z_j = 1) \phi_{z^{j-1}, \epsilon}(x)$.

Define a **term** as follows:

- Every $h \in \mathcal{H}$ is a **term**, and $h$ is said to encode no $\epsilon$.
- For any $\epsilon_1$, let $T_1$ be a **term** that either encodes no $\epsilon$ or encodes only $\epsilon' > \epsilon_1$. Then $T = (1 - \epsilon_1/2)T_1 + (\epsilon_1/2)\Phi_{z,\epsilon_1}$ is a **term** if $|z| \le q(n, 1/\epsilon_1)$. Furthermore, $T$ is said to encode all of the values of $\epsilon$ that $T_1$ encodes plus $\epsilon_1$.

Thus any **term** $T$ may encode up to $n$ values of $\epsilon$, and the values of $\epsilon$ will increase as we get deeper in the **term**. This ensures that all terms have polynomial-sized (in $n$) representations, and the number of terms is finite.

Let $\texttt{list}(T)$ denote the list of all $\epsilon \in S_\epsilon$ that are encoded in $T$. Observe that the definition of term implies that the smallest $\epsilon \in \texttt{list}(T)$ is encoded at the outermost level, and the values increase as we move to the interior. In particular if $\epsilon_1$ is the smallest value in $\texttt{list}(T)$, then $T = (1 - \epsilon_1/2)T_1 + (\epsilon_1/2)\Phi_{z,\epsilon_1}$ for some $z$ and $T_1$ and $\texttt{list}(T_1) = \texttt{list}(T) \setminus \{\epsilon_1\}$. Denote by $\texttt{out}(T)$ the smallest value of $\epsilon$ in $\texttt{list}(T)$ and let $\texttt{next}(T)$ be $T_1$ such that $T = (1 - \texttt{out}(T)/2)T_1 + (\texttt{out}(T)/2)\Phi_{z, \texttt{out}(T)}$.

We consider all terms except those of the form $h \in \mathcal{H}$ to be valid representations. The representation class will also contain more representations, that we shall define shortly.

Consider the following three cases.

(a) The evolutionary algorithm is in a state $T$, such that $\texttt{out}(T) = \epsilon$, where $\epsilon$ is the true parameter of the algorithm. Then (pretending as if $T$ is in $\mathcal{H}$) the results from Section 5 will apply directly. In particular, let $T = r_\epsilon[T_1, z] = (1 - \epsilon/2)T_1 + (\epsilon/2)\Phi_{z,\epsilon}$. Then $\texttt{Neigh}(T, \epsilon) = \{T, r_\epsilon[T, z0], r_\epsilon[T, z1]\}$ if $|z| \le q(n, 1/\epsilon)$. When $|z| = q(n, 1/\epsilon)$, we again define states similar to those in $W_\epsilon$ in 5.3, which allow the algorithm to gradually slide to move to $r_\epsilon[h_z, \sigma]$, $r_\epsilon[T, \sigma]$ or $r_\epsilon[\mathbf{0}, \sigma]$ (but the performance will never go more than $\epsilon$ lower than $T$ with high probability).

(b) The case when $\texttt{out}(T) \le \epsilon$. Let $T_0 = T$, and define $T_i = \texttt{next}(T_{i-1})$ for all $i$, let $k$ be the smallest such that $\texttt{out}(T_k) \ge \epsilon$. (It may happen that $T_k = h$ for some $h \in \mathcal{H}$. Note that,

$$T_1 = (1 - \epsilon_1/2)\Big((1 - \epsilon_2/2)\big(\cdots(1 - \epsilon_{k-1}/2)T_k + (\epsilon_{k-1}/2)\Phi_{z_{k-1}, \epsilon_{k-1}} \cdots\big) + (\epsilon_2/2)\Phi_{z_2, \epsilon_2}\Big) + (\epsilon_1/2)\Phi_{z_1, \epsilon_1} \,.$$

Then since $\epsilon_1 < \epsilon_2 < \cdots < \epsilon_{k-1} \le \epsilon/2$, and every $\epsilon_i$ is a power of 2,

$$\texttt{Perf}_f(T_k, \mathcal{D}) \ge \texttt{Perf}(T_1, \mathcal{D}) - 2(\epsilon_1 + \cdots \epsilon_{k-1}) \ge 4(\epsilon/2) \,.$$

Then $\texttt{Perf}_f(T_k, \mathcal{D}) \ge \texttt{Perf}_f(T, \mathcal{D}) - 2\epsilon$ (because $\epsilon_{k-1} \le \epsilon/2$ ). If $\texttt{out}(T_k) = \epsilon$, let $r_b = T_k$. Otherwise, let $r_b = (1 - \epsilon/2)T_k + (\epsilon/2)\Phi_{\sigma, \epsilon}$. Note that $r_b$ is always a valid **term** (by the above definition) and hence it is in the representation class. Also $\texttt{Perf}_f(r_b, \mathcal{D}) \ge \texttt{Perf}(T, \mathcal{D}) - 3\epsilon$.

(c) The case, when $\texttt{out}(T) > \epsilon$. Let $r_c = (1 - \epsilon/2)T + (\epsilon/2)\Phi_{\sigma,\epsilon}$. Again, $r_c$ is a valid **term** and hence in the representation class. Also in this case $\texttt{Perf}_f(r_c, \mathcal{D}) \ge \texttt{Perf}_f(T, \mathcal{D}) - \epsilon$.

In cases (b) and (c), if we can transition to the representations $r_b$ and $r_c$ respectively, we will have reduced to case (a). However since the moves themselves may be deleterious, we need to add intermediate representations similar to those defined in $W_\epsilon$ in Section 5.3. In particular let $w_0 = T$ (where $T$ may be that of case (b) or (c)). Define $w_k = (1 - k(tu(n, 1/\epsilon)/2))w_0$, where $tu(n, 1/\epsilon)$ is the polynomial upper bound on the tolerances. Define $\texttt{Neigh}(w_k, \epsilon) = \{w_k, w_{k+1}, r_b\}$

(or $\text{Neigh}(w_k, \epsilon) = \{w_k, w_{k+1}, r_c\}$). The idea is the same that the performance reduces gradually until the jump to $r_b$ (or $r_c$) is no longer deleterious. During this time the performance never goes below that of $r_b$ (respectively $r_c$) and hence quasi-monotonicity is maintained. (Although in some cases degradation may be $3\epsilon$, we could just run with higher accuracy (say $\epsilon/4$) to begin with.)

So far we have ignored the drift. However, notice that the number of time steps to get to a representation which encodes the correct value of $\epsilon$ is, with high probability, polynomial (in fact just $2/tu(n, 1/\epsilon)$). Thus by making the drift small enough (though still an inverse polynomial), the function can be made to look essentially unchanging to the evolution algorithm.

## A.8 Proof of Theorem 15

We show that $\text{Neigh}$ is a strictly beneficial neighborhood function for $\mathcal{C}$, $\mathcal{D}$, and $\mathcal{R}$ with $b(n, 1/\epsilon) = \pi^3 n/(2\epsilon)$. The theorem is then an immediate consequence of Theorem 8.

The analysis relies heavily on a couple of useful trigonometric facts. First, it is well known (see, for example, Dasgupta [6]) that under any spherically symmetric distribution $\mathcal{D}$ (for example, the uniform distribution over a sphere), $\text{err}_{\mathcal{D}}(\mathbf{u}, \mathbf{v}) = \arccos(\mathbf{u} \cdot \mathbf{v})/\pi$, where $\arccos(\mathbf{u} \cdot \mathbf{v})$ is the angle between $\mathbf{u}$ and $\mathbf{v}$. We will use this fact repeatedly. We also make use of the following inequalities from Dasgupta et al. [7]. For any $\theta \in [0, \pi/2]$, $2\theta/\pi \leq \sin(\theta) \leq \theta$, and $4\theta^2/\pi^2 \leq 1 - \cos(\theta) \leq \theta^2/2$.

Consider an arbitrary $\mathbf{r} \in \mathcal{R}_n$ and $\mathbf{f} \in \mathcal{C}_n$. To simplify presentation, assume that $r_1 = 1$ and $r_i = 0$ for $i \in \{2, \cdots, n\}$. (Here and for the remainder of this proof, we use the notation $r_i$ and $f_i$ to denote the $i$th components of $\mathbf{r}$ and $\mathbf{f}$, not the values of the representation and ideal function at round $i$ as in previous sections.) This assumption is without loss of generality since we are considering only spherically symmetric distributions. Furthermore, assume that the axes are oriented such that for any $\mathbf{r}' \in \text{Neigh}(\mathbf{r}, \epsilon)$ (except for $\mathbf{r}$ itself), $r'_1 = \cos(\epsilon/(\pi\sqrt{n}))$, $r'_i = \pm \sin(\epsilon/(\pi\sqrt{n}))$ for some $i \in \{2, \cdots, n\}$, and $r'_j = 0$ for all other $j$. This change in basis is also without loss of generality.

Suppose that $\text{Perf}_{\mathbf{f}}(\mathbf{r}, \mathcal{D}) < 1 - \epsilon/2$ since otherwise there is nothing to prove. The condition that we need to prove can be stated as

$$\max_{\mathbf{r}' \in \text{Neigh}(\mathbf{r}, \epsilon)} \text{Perf}_{\mathbf{f}}(\mathbf{r}', \mathcal{D}) \geq \text{Perf}_{\mathbf{f}}(\mathbf{r}, \mathcal{D}) + \frac{1}{b(n, 1/\epsilon)} = \text{Perf}_{\mathbf{f}}(\mathbf{r}, \mathcal{D}) + \frac{2\epsilon}{\pi^3 n} \ .$$

Using the facts that for any unit vectors $\mathbf{u}$ and $\mathbf{v}$, $\text{Perf}_{\mathbf{v}}(\mathbf{u}, \mathcal{D}) = 1 - 2\text{err}(\mathbf{u}, \mathbf{v})$ and $\text{err}_{\mathcal{D}}(\mathbf{u}, \mathbf{v}) = \arccos(\mathbf{u} \cdot \mathbf{v})/\pi$, this condition is equivalent to $\arccos(\max_{\mathbf{r}' \in \text{Neigh}(\mathbf{r}, \epsilon)} \mathbf{r}' \cdot \mathbf{f}) \leq \arccos(\mathbf{r} \cdot \mathbf{f}) - \epsilon/(\pi^2 n)$. By definition of the neighborhood function, there exists a $\mathbf{r}' \in \text{Neigh}(\mathbf{r}, \epsilon)$ such that

$$\mathbf{r}' \cdot \mathbf{f} \geq f_1 \cos\left(\frac{\epsilon}{\pi\sqrt{n}}\right) + \max_{i \in \{2, \cdots, n\}} |f_i| \sin\left(\frac{\epsilon}{\pi\sqrt{n}}\right) \geq f_1 \cos\left(\frac{\epsilon}{\pi\sqrt{n}}\right) + \sqrt{\frac{1 - f_1^2}{n}} \sin\left(\frac{\epsilon}{\pi\sqrt{n}}\right) \ .$$

Using the standard trigonometric equality that for any $\theta$ and $\phi$, $\arccos(\theta) - \arccos(\phi) = \arccos(\theta\phi + \sqrt{(1-\theta^2)(1-\phi^2)})$, we have

$$\arccos(\mathbf{r} \cdot \mathbf{f}) - \frac{\epsilon}{\pi^2 n} = \arccos(f_1) - \arccos\left(\cos\left(\frac{\epsilon}{\pi^2 n}\right)\right)$$

$$= \arccos\left(f_1 \cos\left(\frac{\epsilon}{\pi^2 n}\right) + \sqrt{1 - f_1^2} \sin\left(\frac{\epsilon}{\pi^2 n}\right)\right).$$

Then since $\arccos$ is decreasing in $[0, \pi]$, to prove the result, it is sufficient to show that

$$\arccos\left(f_1 \cos\left(\frac{\epsilon}{\pi\sqrt{n}}\right) + \sqrt{\frac{1-f_1^2}{n}} \sin\left(\frac{\epsilon}{\pi\sqrt{n}}\right)\right) \leq \arccos\left(f_1 \cos\left(\frac{\epsilon}{\pi^2 n}\right) + \sqrt{1 - f_1^2} \sin\left(\frac{\epsilon}{\pi^2 n}\right)\right)$$

or taking the cosine of both sides and rearranging terms,

$$f_1 \left(\cos\left(\frac{\epsilon}{\pi^2 n}\right) - \cos\left(\frac{\epsilon}{\pi\sqrt{n}}\right)\right) \leq \sqrt{1 - f_1^2} \left(\frac{1}{\sqrt{n}} \sin\left(\frac{\epsilon}{\pi\sqrt{n}}\right) - \sin\left(\frac{\epsilon}{\pi^2 n}\right)\right) \ . \quad (7)$$

First consider the case in which $f_1 < 0$. In this case, it is sufficient to show that the difference of cosines on the left hand side of the equation and the difference in sines on the right hand side are both positive. This can be verified easily using the inequalities for sines and cosines given above.

For the rest of this proof, assume that $f_1 > 0$. Since we have assumed that $\text{Perf}_{\mathbf{f}}(\mathbf{r}, \mathcal{D}) < 1 - \epsilon/2$, it follows that $\text{err}(\mathbf{r}, \mathbf{f}) = \arccos(f_1)/\pi > \epsilon/4$, or equivalently, $f_1 < \cos(\epsilon\pi/4)$. Then $\sqrt{1 - f_1^2} > \sqrt{1 - (\cos(\epsilon\pi/4))^2} = \sin(\epsilon\pi/4)$, and for Equation 7 to hold, it is sufficient to show that

$$\cos\left(\frac{\epsilon\pi}{4}\right)\left(\cos\left(\frac{\epsilon}{\pi^2 n}\right) - \cos\left(\frac{\epsilon}{\pi\sqrt{n}}\right)\right) \leq \sin\left(\frac{\epsilon\pi}{4}\right)\left(\frac{1}{\sqrt{n}} \sin\left(\frac{\epsilon}{\pi\sqrt{n}}\right) - \sin\left(\frac{\epsilon}{\pi^2 n}\right)\right) \ . \quad (8)$$

Using the inequalities for sines and cosines given above, we have that

$$\cos\left(\frac{\epsilon\pi}{4}\right)\left(\cos\left(\frac{\epsilon}{\pi^2 n}\right) - \cos\left(\frac{\epsilon}{\pi\sqrt{n}}\right)\right) \le 1\left(1 - \cos\left(\frac{\epsilon}{\pi\sqrt{n}}\right)\right) \le \frac{\epsilon^2}{2\pi^2 n},$$

and

$$\sin\left(\frac{\epsilon\pi}{4}\right)\left(\frac{1}{\sqrt{n}}\sin\left(\frac{\epsilon}{\pi\sqrt{n}}\right) - \sin\left(\frac{\epsilon}{\pi^2 n}\right)\right) \ge \frac{\epsilon}{2}\left(\frac{2\epsilon}{\pi^2 n} - \frac{\epsilon}{\pi^2 n}\right) = \frac{\epsilon^2}{2\pi^2 n}.$$

Therefore Equation 8 holds, Neigh is a strictly beneficial neighbor function, and $\mathcal{C}$ is evolvable with drifting targets. ∎

### A.9 Proof of Theorem 16

We start by analyzing the simpler case in which $\mathcal{D}$ is known to be a spherical Gaussian distribution. In this case, a simpler neighborhood function can be used in which set of "shift" neighbors $N_{\text{sl}}$ is greatly reduced. Below, we explain how to extend this analysis to the case in which $\mathcal{D}$ is an unknown product normal distributions over $\mathbb{R}^n$ and more complex neighborhood function defined in Section 6.2 is used.

Throughout this proof, we use the notation $r_i$ and $f_i$ to denote the $i$th components of $\mathbf{r}$ and $\mathbf{f}$ respectively, and denote by $\mathbf{e}_i$ the basis of $\mathbb{R}^n$. We define the simplified neighborhood function as $\text{Neigh}'(\mathbf{r}, \epsilon) = N_{\text{fl}} \cup N'_{\text{sl}}$, where $N_{\text{fl}} = \{\mathbf{r} - 2r_i\mathbf{e}_i \mid i = 1, \ldots, d\}$ is still the set of representations obtained by flipping the sign of one component of $r$, and $N'_{\text{sl}} = \{\mathbf{r}'_i / \|\mathbf{r}'_i\|_2 \mid \mathbf{r}'_i = \mathbf{r} \pm \beta\mathbf{e}_i, i = 1, \ldots, d\}$ is the set obtained by shifting each component a small amount and renormalizing, with $\beta$ satisfying $\epsilon^2/(6\sqrt{n}) \le \beta \le \epsilon^2/(3\sqrt{n})$.

We first show that for any target function $\mathbf{f}$, increasing $\text{Perf}_{\mathbf{f}}(\mathbf{r}, \mathcal{D})$ is the equivalent to increasing $\mathbf{f} \cdot \mathbf{r}$. The following two lemmas establish this. These lemmas rely on the same trigonometric facts used in the proof of Theorem 15. In particular, under any spherically symmetric distribution $\mathcal{D}$, $\text{err}_{\mathcal{D}}(\mathbf{u}, \mathbf{v}) = \arccos(\mathbf{u} \cdot \mathbf{v})/\pi$, where $\arccos(\mathbf{u} \cdot \mathbf{v})$ is the angle between $\mathbf{u}$ and $\mathbf{v}$, and for any $\theta \in [0, \pi/2]$, $2\theta/\pi \le \sin(\theta) \le \theta$, and $4\theta^2/\pi^2 \le 1 - \cos(\theta) \le \theta^2/2$.

**Lemma 20** *Let $\mathcal{D}$ be a spherical Gaussian distribution. For unit vectors $\mathbf{v}, \mathbf{f}$ and $\alpha \in (0, 1)$, if $\mathbf{f} \cdot \mathbf{v} \ge 1 - \alpha$, then $\text{Perf}_{\mathbf{f}}(\mathbf{v}, \mathcal{D}) \ge 1 - \sqrt{\alpha}$*

**Proof:** Since $\mathbf{f}$ and $\mathbf{v}$ are unit vectors and $\alpha \in (0, 1)$, i.e., $\mathbf{f} \cdot \mathbf{v} > 0$, we may write $\mathbf{f} \cdot \mathbf{v} = \cos(\theta)$ for $\theta \in [0, \pi/2]$. Thus we have that $1 - 4\theta^2/\pi^2 \ge \cos(\theta) \ge 1 - \alpha$, and hence $\theta/\pi \le \sqrt{\alpha}/2$. But $\theta/\pi = \text{err}(\mathbf{f}, \mathbf{v})$ and $\text{Perf}_{\mathbf{f}}(\mathbf{v}, \mathcal{D}) = 1 - 2\text{err}(\mathbf{f}, \mathbf{v}) \ge 1 - \sqrt{\alpha}$. ∎

**Lemma 21** *Let $\mathcal{D}$ be a spherical Gaussian distribution. For unit vectors $\mathbf{u}, \mathbf{v}, \mathbf{f}$, if $\mathbf{f} \cdot \mathbf{u} - \mathbf{f} \cdot \mathbf{v} \ge \omega > 0$, then $\text{Perf}_{\mathbf{f}}(\mathbf{u}, \mathcal{D}) - \text{Perf}_{\mathbf{f}}(\mathbf{v}, \mathcal{D}) \ge \omega/2$.*

**Proof:** Since $\mathbf{u}, \mathbf{v}$ and $\mathbf{f}$ are unit vectors, we may assume that there exit angles $\phi, \theta \in [0, \pi]$ such that $\mathbf{f} \cdot \mathbf{u} = \cos(\phi)$ and $\mathbf{f} \cdot \mathbf{v} = \cos(\theta)$, and that $\phi < \theta$. Since the derivative of the cosine function is lower bounded by $-1$, $\cos(\phi) - \cos(\theta) \le \theta - \phi$. Finally observe that $\text{Perf}_{\mathbf{f}}(\mathbf{u}, \mathcal{D}) - \text{Perf}_{\mathbf{f}}(\mathbf{v}, \mathcal{D}) = 2(\text{err}(\mathbf{f}, \mathbf{v}) - \text{err}(\mathbf{f}, \mathbf{u})) = \frac{2}{\pi}(\theta - \phi) \ge \omega/2$. ∎

The next lemma shows that Neigh$'$ is a strictly beneficial neighborhood function. More than the lemma statement itself, it is the analysis of this lemma that is important to us. Below we will show how to extend this analysis to the case in which $\mathcal{D}$ is a product normal distribution.

**Lemma 22** *Let $\mathcal{C}$ be the class of homogeneous linear separators, $\mathcal{R}$ be the class of homogeneous linear separators represented by unit length normal vectors, and $\mathcal{D}$ be a spherical Gaussian distribution. Define Neigh$'$ as in the previous paragraph and let $p$ be any polynomial such that $p(n, 1/\epsilon) \ge 3n$. Then Neigh$'$ is a strictly beneficial neighborhood function for $\mathcal{C}$, $\mathcal{D}$, and $\mathcal{R}$, with $b(n, 1/\epsilon) = 144n/\epsilon^6$.*

**Proof:** Let $\rho = \epsilon^3/(12\sqrt{n})$ and $\eta = \epsilon^2/(3\sqrt{n})$. By assumption, $\beta$ then satisfies $\eta/2 \le \beta \le \eta$.

Consider an arbitrary $\mathbf{r} \in \mathcal{R}_n$ and $\mathbf{f} \in \mathcal{C}_n$. If there exists $\mathbf{r}' \in N_{\text{fl}}$ such that $\text{Perf}_{\mathbf{f}}(\mathbf{r}', \mathcal{D}) - \text{Perf}_{\mathbf{f}}(\mathbf{r}, \mathcal{D}) \ge 1/b(n, 1/\epsilon)$, then we are done, so assume that there is no element in $N_{\text{fl}}$ with this property. In this case, we then claim that one of the following must hold for all $i = 1, \ldots, n$: (i) $r_i$ and $f_i$ have the same sign, (ii) $|r_i| \le \rho$, or (iii) $|f_i| \le \rho$. If none of these properties hold, then $\mathbf{f} \cdot (\mathbf{r} - 2r_i\mathbf{e}_i) - \mathbf{f} \cdot \mathbf{r} = -2r_i f_i \ge 2\rho^2$ and by Lemma 21, the change in performance is at least $\rho^2 = 1/b(n, 1/\epsilon)$.

In the rest of the analysis we assume that if no flip is a beneficial mutation (by at least $1/b(n, 1/\epsilon)$), all $r_i$ and $f_i$ are in the interval $[-\rho, 1]$. The reason this does not affect generality is this: Suppose one of them is smaller than $-\rho$, we know that the other one then lies in the interval $[-1, \rho]$. We can now assume that the basis we were working with actually contained $-\mathbf{e}_i$ rather than $\mathbf{e}_i$. (This is useful for analysis, so that we can only consider mutations which increase the value of any component.) However, the neighborhood contains both mutations. Thus, we may assume that all $f_i$ and $r_i$ are in $[-\rho, 1]$, and hence $\mathbf{f} \cdot \mathbf{r} \geq -\rho(\|\mathbf{f}\|_1 + \|\mathbf{r}\|_1) \geq -2\rho\sqrt{n}$.

In this situation if there is no $i$ for which $r_i \leq f_i - \eta$, then $\mathbf{f} \cdot \mathbf{r} \geq 1 - \epsilon^2$, and we are already close to optimal. (as shown below)

$$\mathbf{f} \cdot \mathbf{r} = \sum_i f_i r_i = \sum_{\substack{i \\ f_i \in [-\rho, 0)}} f_i r_i + \sum_{\substack{i \\ f_i \in [0,1]}} f_i r_i \geq -\rho\|\mathbf{r}\|_1 + \sum_{\substack{i \\ f_i \in [0,1]}} f_i(f_i - \eta)$$

$$\geq \sum_{\substack{i \\ f_i \in [0,1]}} f_i^2 - \rho\|\mathbf{r}\|_1 - \eta\|\mathbf{f}\|_1 \geq 1 - n\rho^2 - \sqrt{n}\rho - \sqrt{n}\eta$$

Suppose there exists an $i$ for which $r_i \leq f_i - \eta$, then $f_i \geq \eta - \rho \geq \eta/2 > 0$. Let $\mathbf{r}' = \mathbf{r} + \beta\mathbf{e}_i$, with $\eta/2 \leq \beta \leq \eta$. Then $\|\mathbf{r}'\|_2 = \sqrt{1 + 2\beta r_i + \beta^2}$. From elementary algebra, we get the inequality that $1 + \beta r_i \leq \sqrt{1 + 2\beta r_i + \beta^2} \leq 1 + \beta r_i + \beta^2/2$ (assuming $\beta r_i \in (-1, 1)$, which is true). Then consider the following quantity of interest:

$$\mathbf{f} \cdot \frac{\mathbf{r}'}{\|\mathbf{r}'\|_2} - \mathbf{f} \cdot \mathbf{r} = \frac{\mathbf{f} \cdot \mathbf{r}' - \sqrt{1 + \beta r_i + \beta^2}(\mathbf{f} \cdot \mathbf{r})}{\|\mathbf{r}'\|_2}$$

Since $1/2 \leq \|\mathbf{r}'\|_2 \leq 2$, if the quantity in the numerator is positive (as we will show) we have

$$2\left(\mathbf{f} \cdot \frac{\mathbf{r}'}{\|\mathbf{r}'\|_2} - \mathbf{f} \cdot \mathbf{r}\right) \geq \mathbf{f} \cdot (\mathbf{r} + \beta\mathbf{e}_i) - \sqrt{1 + 2\beta r_i + \beta^2}(\mathbf{f} \cdot \mathbf{r})$$

$$= \beta f_i + \mathbf{f} \cdot \mathbf{r}\left(1 - \sqrt{1 + 2\beta r_i + \beta^2}\right) . \tag{9}$$

Notice that by our setting of parameters, $r_i \geq -\rho \geq -\beta/2$, thus the quantity under the square root sign is greater than 1. When $\mathbf{f} \cdot \mathbf{r} < 0$, the second term in the above expression is actually positive, and hence the total quantity is at least as much as the first term which is at least $\beta\eta/2 = \eta^2/4 \geq 2/b(n, 1/\epsilon)$. Thus we will consider the case when $\mathbf{f} \cdot \mathbf{r} \geq 0$. In that case continuing from equation (9) and using the fact that $\sqrt{1 + 2\beta r_i + \beta^2} \leq 1 + \beta r_i + \beta^2/2$, we get

$$\mathbf{f} \cdot \frac{\mathbf{r}'}{\|\mathbf{r}'\|_2} - \mathbf{f} \cdot \mathbf{r} \geq \frac{1}{2}\left(\beta f_i - (1 - \epsilon^2)(\beta r_i + \beta^2/2)\right) .$$

Since $r_i + \beta/2 \leq r_i + \eta \leq f_i$, this is greater than $\beta f_i \epsilon^2/2 \geq 2/b(n, 1/\epsilon)$. Hence `Neigh'` is a strictly beneficial neighborhood function. ∎

**Product Normal Distributions**

We now describe how the analysis above can be adjusted to handle product normal distributions with polynomial variances. Recall that $\sigma_1, \ldots, \sigma_n$ are the standard deviations of the distribution $\mathcal{D}$ in each of the $n$ dimensions, and that $1 \geq \sigma_i \geq (1/n)^k$ for all $i$ for some constant $k$ (which is known by the algorithm). Assume without loss of generality that $1 = \sigma_1 \geq \sigma_2 \geq \cdots \geq \sigma_n \geq (1/n)^k$.

Define $\tau(x_1, \ldots, x_n) = (x_1/\sigma_1, \cdots, x_n/\sigma_n)$ and $\lambda(x_1, \ldots, x_n) = (\sigma_1 x_1, \cdots, \sigma_n x_n)$. Note that the transformations $\tau$ and $\lambda$ are inverses. Let $\mathcal{N}[\mathbf{0}, \Sigma]$ denote the distribution with covariance matrix $\Sigma = \text{diag}(\sigma_1^2, \ldots, \sigma_n^2)$, and let $\mathcal{N}[\mathbf{0}, \mathrm{I}]$ denote the spherical normal distribution with variance 1. Note that if $\mathbf{x}$ is distributed according to $\mathcal{N}[\mathbf{0}, \Sigma]$, $\tau(\mathbf{x})$ is distributed according to $\mathcal{N}[\mathbf{0}, \mathrm{I}]$. For any vector $\mathbf{f}$ and $\mathbf{r}$, we have

$$\texttt{err}_{\mathcal{N}[\mathbf{0}, \Sigma]}(\mathbf{f}, \mathbf{r}) = \Pr_{\mathbf{x} \sim \mathcal{N}[\mathbf{0}, \Sigma]}[\text{sign}(\lambda(\mathbf{f}) \cdot \tau(\mathbf{x})) \neq \text{sign}(\lambda(\mathbf{r}) \cdot \tau(\mathbf{x})]$$

$$= \Pr_{\mathbf{z} \sim \mathcal{N}[\mathbf{0}, \mathrm{I}]}[\text{sign}(\lambda(\mathbf{f}) \cdot \mathbf{z}) \neq \text{sign}(\lambda(\mathbf{r}) \cdot \mathbf{z})] = \texttt{err}_{\mathcal{N}[\mathbf{0}, \mathrm{I}]}(\lambda(\mathbf{f}), \lambda(\mathbf{r})) .$$

We assume that $\|\mathbf{f}\|_2 = 1$ and that our representations consist of vectors also of unit norm. Then, for all $\mathbf{r}$ we have, $(1/n)^k \leq \|\lambda(\mathbf{r})\|_2 \leq 1$. Observe that the "flips" are invariant with respect to $\lambda$, i.e., $\lambda(\mathbf{r} - 2r_i \mathbf{e}_i) = \lambda(\mathbf{r}) + 2\lambda(\mathbf{r})_i \mathbf{e}_i$. Because of this, we can consider the same set $N_{\text{fl}}$ of flips as in the spherical distribution case.

Unfortunately, it does not suffice to use the same set of "shift" mutations $N'_{\text{sl}}$. Let $\mathbf{r}$ be our current representation and let $\mathbf{r}' = \mathbf{r} + \gamma \mathbf{e}_i$. Consider the two vectors $\lambda(\mathbf{r})/\|\lambda(\mathbf{r})\|_2$ and $\lambda(\mathbf{r}')/\|\lambda(\mathbf{r})\|_2$, and consider their $i$th components, which are $\sigma_i r_i / \|\lambda(\mathbf{r})\|_2$ and $\sigma_i(r_i + \gamma)/\|\lambda(\mathbf{r})\|_2$ respectively. The difference between the two is $\sigma_i \gamma / \|\lambda(\mathbf{r})\|_2$. As in the proof of Lemma 22, let $\eta = \epsilon^2/(3\sqrt{n})$. If $\gamma$ took a certain value such that $\eta/2 \leq \sigma_i \gamma / \|\lambda(\mathbf{r})\|_2 \leq \eta$, then by the same analysis in the proof of Lemma 22, this would be a beneficial mutation.

Since we don't know the values of $\sigma_i$ and $\|\lambda(\mathbf{r})\|_2$, we use the following trick: Let

$$N_i = \left\{ \mathbf{r} \pm \left(\frac{j\eta}{4n^k}\right) \mathbf{e}_i \mid 1 \leq j \leq 4n^k \right\}.$$

Now consider the quantity

$$\gamma_j = \frac{\sigma_i}{\|\lambda(\mathbf{r})\|_2} \frac{\eta j}{4n^k}.$$

Observe that $(1/n)^k \leq \sigma_i/\|\lambda(\mathbf{r})\|_2 \leq n^k$. Thus $\gamma_1 \leq \eta/4$, $\gamma_{4n^{2k}} \geq \eta$, and finally $\gamma_j - \gamma_{j-1} \leq \eta/4$; at least one $j$ satisfies $\eta/2 \leq \gamma_j \leq \eta$. Let $N_{\text{sl}} = \{\mathbf{r}'/\|\mathbf{r}'\|_2 | \mathbf{r}' \in N_i, i \in \{1, \ldots, n\}\}$. (This is the same set $N_{\text{sl}}$ defined in Section 6.2, only in slightly different notation.) With $\text{Neigh}(\mathbf{r}, \epsilon) = N_{\text{fl}} \cup N_{\text{sl}}$, Neigh is then a strictly beneficial neighborhood function with respect to any product normal distribution with variance lower bounded by $(1/n)^k$ as desired. The benefit polynomial remains the same as in Lemma 22, $b(n, 1/\epsilon) = 144n/\epsilon^6$, though the neighborhood size is larger. ∎

### A.10 Proof of Theorem 17

We show that Neigh is a strictly beneficial neighborhood function for $\mathcal{C}$, $\mathcal{D}$, and $\mathcal{R}$ with benefit polynomial $b(n, 1/\epsilon) = 9/\epsilon^2$. The theorem is then an immediate consequence of Theorem 8.

Since $q = \lceil \log_2(3/\epsilon) \rceil$, it follows that $\epsilon/6 \leq 2^{-q} \leq \epsilon/3$. We make use of this repeatedly.

Consider an arbitrary $r \in \mathcal{R}_n$ and $f \in \mathcal{C}_n$. As in Diochnos and Turán [8], we define $m$ to be the number of "mutual" variables shared between $f$ and $r$, $u$ to be the number of "undiscovered" variables that appear in $f$ but not $r$, and $w$ to be the number of "wrong" variables that appear in $r$ but not $f$. Thus $|f| = m + u$ and $|r| = m + w$. The functions $r$ and $f$ disagree if and only if all $m$ mutual variables are true and either all $u$ undiscovered variables are true while some wrong variable is false, or all $w$ wrong variables are true while some undiscovered variable is false. Therefore if $\mathcal{D}$ is uniform, $\text{err}_\mathcal{D}(f, r) = 2^{-m}(2^{-u}(1 - 2^{-w}) + 2^{-w}(1 - 2^{-u})) = 2^{-m-u} + 2^{-m-w} - 2^{1-m-u-w}$, so

$$\text{Perf}_f(r, \mathcal{D}) = 1 - 2^{1-m-u} - 2^{1-m-w} + 2^{2-m-u-w} = 1 - 2^{1-|f|} - 2^{1-|r|} + 2^{2-m-u-w}. \quad (10)$$

We start by considering the case in which the target is "long", that is, $|f| = m + u \geq q + 1$. If $|r| = m + w = q$, then $\text{Perf}_f(r, \mathcal{D}) > 1 - 2^{1-|f|} - 2^{1-|r|} \geq 1 - 2^{-q} - 2^{1-q} = 1 - 3 \cdot 2^{-q} \geq 1 - \epsilon$, and the performance of $r$ with respect to $f$ is already good enough. This is because both $f$ and $r$ are almost always false under the uniform distribution. On the other hand, if $|r| < q$, then there must exist a neighbor $r'$ in the set $\mathcal{N}^+(r)$ such that the variable contained in $r'$ but not $r$ is an undiscovered variable of $f$. Then $\text{Perf}_f(r', \mathcal{D}) - \text{Perf}_f(r, \mathcal{D}) = 2^{-|r|} > 2^{-q} = \epsilon/6$.

Now consider the case in which the target is "short", that is, $f = m + u \leq q$. Suppose that $u = 0$ (so the variables in $f$ are a strict subset of the variables in $r$). If $w = 0$, then $f$ and $r$ must be identical, so assume $w > 0$. Then there must exist a neighbor $r'$ in the set $\mathcal{N}^-$ such that the variable contained in $r$ but not $r'$ is a wrong variable. From Equation 10, for this neighbor $r'$, $\text{Perf}_f(r', \mathcal{D}) - \text{Perf}_f(r, \mathcal{D}) = 2^{1-|r|} \geq 2^{1-q} \geq \epsilon/3$. On the other hand, suppose that $u > 0$. If $|r| = m + w < q$, then there must exist a neighbor $r'$ in the set $\mathcal{N}^+(r)$ such that the variable contained in $r'$ but not $r$ is an undiscovered variable of $f$. As above, $\text{Perf}_f(r', \mathcal{D}) - \text{Perf}_f(r, \mathcal{D}) = 2^{-|r|} > 2^{-q} = \epsilon/6$. Finally, if $|r| = m + w = q$, then there must exist a neighbor $r'$ in the set $N^{\pm}$ such that the variable contained in $r$ but not $r'$ is wrong and the variable contained in $r'$ but not $r$ is an undiscovered variable of $f$. In this case, from Equation 10, $\text{Perf}_f(r', \mathcal{D}) - \text{Perf}_f(r, \mathcal{D}) = 2^{2-m-u-w} \geq 2^{2-2q} \geq \epsilon^2/9$.

We have shown that whenever $\text{Perf}_f(r, \mathcal{D}) < 1 - \epsilon$, there exists an $r' \in \text{Neigh}(r, \epsilon)$ such that $\text{Perf}_f(r', \mathcal{D}) - \text{Perf}_f(r, \mathcal{D}) \geq \epsilon^2/9$, so Neigh is a strictly beneficial neighborhood function. ∎

### A.11 Proof of Theorem 18

We show that Neigh is a strictly beneficial neighborhood function for $\mathcal{C}$, $\mathcal{D}$, and $\mathcal{R}$ with benefit polynomial $b(n, 1/\epsilon) = 9/\epsilon^2$. The theorem is then an immediate consequence of Theorem 8.

Consider an arbitrary $r \in \mathcal{R}_n$ and $f \in \mathcal{C}_n$. As in the proof of Theorem 17, we start by considering the case in which the target is "long", that is, $|f| \geq q + 1$, where $q = \lceil \log_2(3/\epsilon) \rceil$. If $|r| = q$, then as before, $\text{Perf}_f(r, \mathcal{D}) > 1 - 2^{1-|f|} - 2^{1-|r|} \geq 1 - \epsilon$, and the performance of $r$ with respect to $f$ is already good enough. If $r$ does not contain a literal that is a negation of a literal in $f$, then Equation 10 holds and just as in the proof of Theorem 17, there must exist a neighbor $r'$ in the set $\mathcal{N}^+(r)$ such that $\texttt{Perf}_f(r', \mathcal{D}) - \texttt{Perf}_f(r, \mathcal{D}) = 2^{-|r|} > 2^{-q} = \epsilon/6$. On the other hand, if $r$ does contain at least one literal that is a negation of a literal in $f$, then $f$ and $r$ are never simultaneously true and so $\texttt{Perf}_f(r, \mathcal{D}) = 1 - 2(2^{-|f|} + 2^{-|r|}) = 1 - 2^{1-|f|} - 2^{1-|r|}$. In this case, by a similar argument, a neighbor $r' \in \mathcal{N}^+$ has performance $1 - 2^{1-|f|} - 2^{-|r|}$, so $\texttt{Perf}_f(r', \mathcal{D}) - \texttt{Perf}_f(r, \mathcal{D}) = 2^{-|r|} > 2^{-q} \geq \epsilon/6$.

Now consider the case in which $f$ is "short". If $r$ does not contain a literal that is a negation of a literal in $f$, then Equation 10 holds and the case-by-case analysis is identical to the analysis in the proof of Theorem 17. Suppose $r$ contains at least one literal that is the negation of a literal in $f$. In this case, as above, $\texttt{Perf}_f(r, \mathcal{D}) = 1 - 2^{1-|f|} - 2^{1-|r|}$. Let $r' \in \mathcal{N}'(r)$ be the conjunction obtained by starting with $r$ and negating all literals in $S$. From Equation 10, we have that $\texttt{Perf}_f(r', \mathcal{D}) \geq 1 - 2^{1-|f|} - 2^{1-|r|} + 2^{2-|f|-|r|}$, and so $\texttt{Perf}_f(r', \mathcal{D}) - \texttt{Perf}_f(r, \mathcal{D}) \geq 2^{2-|f|-|r|} \geq 2^{2-2q} \geq \epsilon^2/9$. The lemma statement follows. ∎